%% file: main.tex
\documentclass[11pt]{article}
\usepackage{b3}   %

{\fontfamily{lmr}\selectfont}%
\DeclareFontShape{T1}{lmr}{bx}{sc}{<->ssub*lmr/m/sc}{}

\input{math_commands.tex}

\usepackage{xspace}
\usepackage{multirow}
\usepackage{array}
\usepackage{flafter}
\usepackage{placeins}
\usepackage{needspace}
\usepackage{pifont}      %
\usepackage{colortbl}    %
\graphicspath{{plots/}}

\let\geometry\relax
\input{paper_macros.tex}

\AddToHook{begindocument/end}{\global\let\geometry\posemodelmacro}

\title{Never Look Back}
\jhusubtitle{Understanding Persistence in 3D Object Memory from Egocentric Videos}
\jhurunningtitle{Never Look Back}
\hypersetup{pdftitle={Never Look Back: Understanding Persistence in 3D Object Memory from Egocentric Videos}}

\hypersetup{pdfauthor={Shravan S Chaudhari, William Paul, Suchi Saria, Rama Chellappa, Homanga Bharadhwaj}}

\author{Shravan~S~Chaudhari\textsuperscript{1} \and William~Paul\textsuperscript{2}
  \and Suchi~Saria\textsuperscript{1}\\[0.2em]
  Rama~Chellappa\textsuperscript{1*} \and Homanga~Bharadhwaj\textsuperscript{1*}}
\jhuinstitution{\textsuperscript{1}Johns Hopkins University\qquad
  \textsuperscript{2}Johns Hopkins University Applied Physics Laboratory}
\jhucontact{\textsuperscript{*}Equal advising}
\jhuprojecturl{ledger-3d.github.io}
\jhucodeurl{github.com/LEDGER-3D/LEDGER}

\begin{document}

\maketitle
\addtocontents{toc}{\protect\setcounter{tocdepth}{-1}}

\input{sections/teaser_figure.tex}
\begin{abstract}
As we move through the world and carry out everyday tasks, we encounter objects that may become relevant only later. We are capable of recalling where we left something or what was inside a container, even without knowing we would need it later. Here, we study how an embodied assistant can build a similar memory from egocentric videos, by observing a person's day-to-day activities. We present \sysname{}, a persistent 3D object memory that combines object locations, their histories, and contextual descriptions. It associates observations across the recording and retains objects after they leave the view, including those the person never touches. It clusters each object's observations by resting locations and records a move only after repeated evidence, reducing the effect of localization noise. Short descriptions preserve details such as an object's contents or supporting surface. It saves these records to later answer spatial questions without having to access the original images or video.
Our memory raises \hdepic{} accuracy from 29.7\% to 42.6\%, \ucs{} accuracy from
33.8\% to 38.5\% and localizes \egofour{} objects with a 0.99\,m median error
on returned predictions. Our analyses identify complementary
roles for temporal persistence, contextual descriptions, and retrieval.
Our study on 100 stitched streams of multiple scenes each further exposes failures in both
retrieval and construction. Per-scene construction partially
recovers the performance lost across scene changes compared to that of single scene streams.
\end{abstract}

\input{sections/introduction.tex}
\input{sections/related_work.tex}
\input{sections/method.tex}
\input{sections/experiments.tex}

\clearpage
\bibliography{iclr2027_conference}
\bibliographystyle{plainnat}

\clearpage
\appendix
\input{sections/appendix_index.tex}      %
\input{sections/appendix_method.tex}     %
\input{sections/appendix_hd_epic.tex}    %
\input{sections/appendix_vq3d.tex}       %
\input{sections/appendix_ucs.tex}        %
\input{sections/appendix_streams.tex}    %
\input{sections/appendix_length.tex}     %
\input{sections/appendix_analysis.tex}   %
\input{sections/appendix_cost.tex}       %

\end{document}

%% file: math_commands.tex
\usepackage{amsmath,amsfonts,bm}

\def\eqref#1{equation~\ref{#1}}

\def\1{\bm{1}}

\DeclareMathAlphabet{\mathsfit}{\encodingdefault}{\sfdefault}{m}{sl}
\SetMathAlphabet{\mathsfit}{bold}{\encodingdefault}{\sfdefault}{bx}{n}

%% file: paper_macros.tex
\definecolor{vidred}{RGB}{190,30,30}
\definecolor{vidgreen}{RGB}{20,120,50}
\DeclareRobustCommand{\vidyes}{\textcolor{vidred}{\textbf{yes}}}
\DeclareRobustCommand{\vidno}{\textcolor{vidgreen}{\textbf{no}}}
\DeclareRobustCommand{\vidsq}{\textcolor{vidred}{\rule{0.85ex}{0.85ex}}\,}
\DeclareRobustCommand{\novidsq}{\textcolor{vidgreen}{\rule{0.85ex}{0.85ex}}\,}

\newcommand{\sysnameagent}{\textsc{Ledger}$_{\text{agent}}$\xspace}
\newcommand{\sysname}{\textsc{Ledger}\xspace}

\newcommand{\hdepic}{HD-EPIC\xspace}
\newcommand{\catperception}{3D Perception\xspace}
\newcommand{\catmotion}{Object Motion\xspace}
\newcommand{\egofour}{Ego4D\xspace}
\newcommand{\vqthreed}{VQ3D\xspace}
\newcommand{\teleego}{TeleEgo\xspace}
\newcommand{\ucs}{UCS-Bench\xspace}

\newcommand{\gpt}{GPT-5.4\xspace}
\newcommand{\gptshort}{GPT\xspace}
\newcommand{\qwen}{Qwen3.5-9B\xspace}
\newcommand{\qwenshort}{Qwen\xspace}

\newcommand{\tagger}{Qwen3.5-9B\xspace}
\newcommand{\detector}{YOLO-World\xspace}

\newcommand{\detectorfull}{YOLO-World~\citep{cheng2024yoloworld}\xspace}

\newcommand{\segmenter}{SAM~3\xspace}

\newcommand{\segmenterfull}{SAM~3~\citep{carion2025sam3}\xspace}

\newcommand{\trackerfull}{the SAM~3 video tracker~\citep{carion2025sam3}\xspace}

\newcommand{\lifter}{WildDet3D\xspace}

\newcommand{\lifterfull}{WildDet3D~\citep{huang2026wilddet3dscalingpromptable3d}\xspace}

\newcommand{\geometry}{FastVGGT\xspace}

\newcommand{\geometryfull}{FastVGGT~\citep{shen2025fastvggt}\xspace}
\newcommand{\geometrybase}{VGGT~\citep{wang2025vggt}\xspace}
\newcommand{\depthanything}{Depth Anything~3\xspace}
\newcommand{\egoloc}{EgoLoc\xspace}

\makeatletter
\@ifundefined{paperdraft}{
  \newcommand{\todo}[1]{}
  \newcommand{\note}[1]{}
  
}{
  \newcommand{\todo}[1]{\textcolor{red}{[TODO: #1]}}
  \newcommand{\note}[1]{\textcolor{blue}{[NOTE: #1]}}
  
}
\makeatother

\newcommand{\cmark}{\textcolor{green!50!black}{\ding{51}}}
\newcommand{\xmark}{\textcolor{red!70!black}{\ding{55}}}
\newcommand{\pmark}{\textcolor{orange!80!black}{$\circ$}}   %
\newcommand{\na}{n/a}

\makeatletter
\newcommand{\rqlabel}[2]{\phantomsection\def\@currentlabel{#1}\label{#2}}
\makeatother

%% file: sections/teaser_figure.tex
\begingroup
\setlength{\intextsep}{4pt plus 2pt minus 2pt}%
\begin{figure}[h!]
\centering
\includegraphics[width=0.95\textwidth]{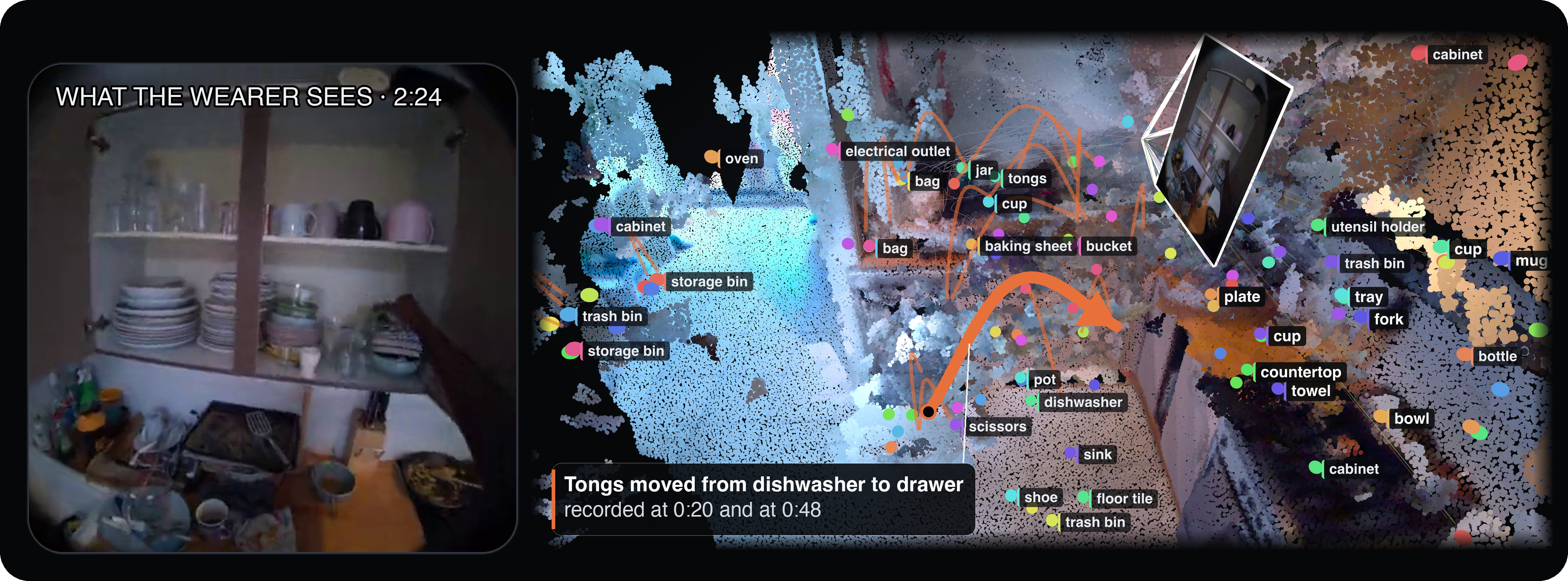}\\[2pt]
\includegraphics[width=0.95\textwidth]{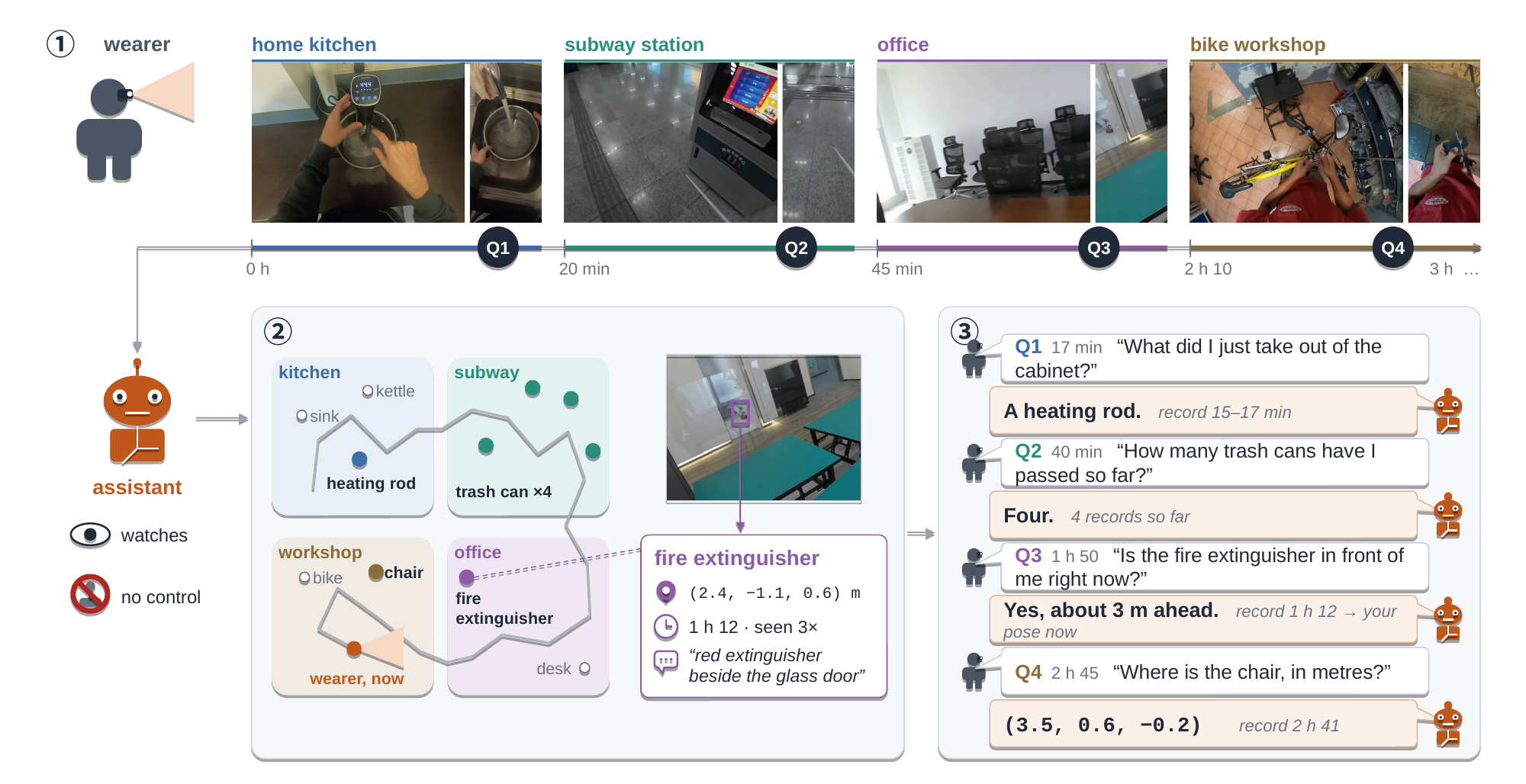}
\vspace*{-0.25cm}
\caption{\textbf{Persistent object memory from egocentric video.} \textbf{Top:} \sysname's
memory 2\,min 24\,s into a scene in the kitchen: tracked objects (bubbles) persist after leaving
view, arcs show recorded moves (bold: the tongs), and the white frame is the wearer's camera, with its current
view inset; the point cloud is for display only. \textbf{Bottom:} \textcircled{\scriptsize 1}~A wearer
or an assistant (physical or digital) works for hours across scenes while questions arrive.
\textcircled{\scriptsize 2}~It keeps a record for each object, with its 3D position, times seen,
and a short description, on a map of its path. \textcircled{\scriptsize 3}~Later questions are
answered from these records alone, without replaying the video.}
\label{fig:teaser}
\end{figure}
\endgroup

%% file: sections/introduction.tex
\section{Introduction}
\label{sec:intro}
Humans typically perform several tasks during the day like cooking, clearing out
the garage, playing football, or other diverse activities. We are reliably good at tracking our environments
and our positions relative to the objects of the scene we are in \citep{wang2000updating,burgess2006spatial}.
Even after exiting the scene, we can still answer certain questions about it like where
we left the wrench, which shelf had the spare batteries, or where the serving bowl was last placed \citep{postma2008objectlocation,hollingworth2002accurate}.
Fundamentally, we keep a running memory of where potentially important objects are
\citep{land1999roles,helbing2020search}, freeing us from having to exactly retrace our steps in either our memories
or the scene itself \citep{tolman1948cognitive,epstein2017cognitivemap}.
Similarly, in long horizon explorations spanning multiple scenes, we are capable of both tracking
objects across scenes \citep{warren2017wormholes} and revising our memory for familiar objects displaced when we were absent from a scene \citep{hollingworth2005online,hupbach2008dynamics}.
Our goal is to develop embodied spatial memory that supports recall of
previously visited scenes and incorporates new observations of objects and
changes in familiar environments. A natural setting to study these capabilities
is long egocentric videos about everyday human
activity (Figure~\ref{fig:teaser}). Such videos capture
people moving through diverse scenes, interacting with objects, and revisiting
places over time. This enables us to study what an embodied agent can retain
from a person's visual experience without controlling where that person looks
or what they do.

Building such a memory from these recordings is challenging. Objects can be occluded, briefly visible or captured from an alternate
viewpoint~\citep{grauman2022ego4d,Damen2022RESCALING,tang2023egotracks}. They may also
move between observations, and errors in 3D localization can make an agent mistake a stationary object
for a moving one. Moreover, the agent does not know during
exploration which objects or events will be relevant to a question asked later. It should decide
what information to retain from these incomplete observations. This includes information about
objects that are only seen in passing. In this work, we study four aspects of this problem: which objects are recorded
in the memory, how to record changes in their locations, what information is retained beyond
their 3D positions, and how earlier observations can be retrieved as more scenes are
added. Our goal here is to understand how these design choices impact the spatial questions
an agent can answer after exploration, without revisiting the scene or replaying the video.

\input{sections/capability_table.tex}  %
Recent works have developed object memories that retain identities and locations after objects
leave the view~\citep{Plizzari2025OSNOM,chalk2026whareformer,goletto2024amego,manigrasso2024esom}.
Object-centric maps and scene graphs organize observations through the spatial relationships between
objects~\citep{gu2024conceptgraphs,schmid2024khronos}. Language-model agents, in contrast, write their
observations into a structured record of objects, places, and their states, which the model later reads to
answer questions~\citep{anwar2024remembr,fan2025embodied,yu2026memora,wang2026ucsbench}.
Another line of work simply uses video-language models to answer from frames
sampled from the recording at question time without explicitly maintaining a record of the scene~\citep{cheng2024videollama2,zhang2024longva,zhang2024llavavideo,
geminiteam2024gemini15}, including models trained particularly for spatial
reasoning~\citep{wu2025spatialmllm,yang2025cambrians}. Streaming models compress past frames in a
memory of visual tokens~\citep{zhang2025flashvstream,qian2025dispider,huang2025videochatonline}.
Visual query localization methods search the recording again for every
query~\citep{grauman2022ego4d,mai2023egoloc}. Table~\ref{tab:capabilities} compares what these
memories keep and what they need at query time.
To guide memory design, we need to decide which retained evidence supports later questions and to
distinguish failures of representation from retrieval and reasoning failures.

For example, restricting memory to objects involved in hand
interactions~\citep{goletto2024amego,Plizzari2025OSNOM} can exclude visible objects that still affect the agent or are relevant later on.
Although retaining an object's position may support a location question, it does not
preserve its contents or its movement history. Even a sequence of positions can mislead when
localization noise is recorded as movement. Further failures can arise across scenes when
storing observations into memory or while retrieving relevant records. These examples
motivate us to address our core question: what must an agent preserve from past observations to answer spatial
questions it could not anticipate?

To this end, we present \sysname{} (Long-horizon Egocentric Descriptions, Geometry, and Event Records), a persistent 3D object memory that combines geometric
observations with contextual descriptions. We construct the memory offline from sampled
egocentric recordings, using supplied or estimated camera poses. We detect objects with an
open-vocabulary detector, associate their observations over time, store them whether or not the
wearer interacts with them, and keep their records after they leave view. We group each object's timestamped
3D observations into \emph{rest segments}. Each rest segment for an object represents its observations close to the same
resting location. A new segment indicates that the object has moved, which is only recorded when multiple subsequent
observations support it. We use a persistence rule to reduce false motion detections caused
by localization noise.
We add a short contextual description to each segment to capture details such as what a
container held or which surface it rested on.
We first identify the objects referred to in the question and retrieve their records. A language model then answers using these records as text, without seeing images or the original video. This tests whether the stored locations, histories, and descriptions provide enough information to answer questions about objects no longer in view.

We evaluate \sysname{} on all 2,400 questions in HD-EPIC's 3D Perception and
Object Motion categories~\citep{perrett2025hdepic}, named-object localization on
Ego4D VQ3D~\citep{grauman2022ego4d}, and 2,771 timestamped questions on
UCS-Bench~\citep{wang2026ucsbench}. On HD-EPIC, adding our memory raises GPT-5.4's%
\footnote{\gpt runs without extended reasoning (thinking) in all our experiments.}
prototype-averaged accuracy from $29.7\%$ to $42.6\%$.
On UCS-Bench, the answerer sees 48 frames. Adding \sysname{} yields
$46.3\%$ accuracy, compared with $42.9\%$ when adding DirectMe's graph.
On VQ3D, our memory returns object locations with a median error of
0.99\,m among the queries for which it returns a prediction.

\paragraph{Contributions.}
\begin{itemize}\setlength{\itemsep}{1pt}\setlength{\parskip}{0pt}
    \item \textbf{A reusable 3D memory of past observations.}
    We introduce \sysname{}, a persistent object memory built offline
    from egocentric videos with supplied or estimated poses, retaining
    locations, movement histories, and contextual descriptions, even for
    untouched objects.

    \item \textbf{Spatial answering and localization across benchmarks.}
    We demonstrate improved text-only question answering on \hdepic{},
    named-object localization on \egofour{} \vqthreed{}, and gains over
    frame-only answering on \ucs{}, and compare against rebuilt object,
    caption, and graph memories to show where our representation helps and
    what it misses.

    \item \textbf{Evidence for how to build and read memory.}
    Ablations isolate temporal persistence, contextual descriptions, and
    geometry, and shared readers reveal how retrieval shapes a
    representation's apparent advantage; costlier iterative search does not
    improve every task.

    \item \textbf{An analysis of accumulating experience.}
    Across recording lengths and 100 paired stitched streams, memory
    retains its gains up to 75 minutes, while scene changes expose failures
    in construction and retrieval that building memory per scene partially
    repairs.
\end{itemize}

%% file: sections/capability_table.tex
\begin{table}[t]
\centering
\scriptsize
\setlength{\tabcolsep}{1.5pt}
\renewcommand{\arraystretch}{1.1}
\caption{What existing memories built from egocentric or embodied video keep, and what
they need at query time. \cmark: supported and demonstrated; \pmark: partly or only under
conditions; \xmark: not supported; N.A.: not applicable. The memories differ in their objects'
coverage, tracking of motion history, input conditions for construction or
handling of scene changes, and query time support for no frame access, language queries and 3D location requests.}
\label{tab:capabilities}
\begin{tabular}{l cccc ccc ccc}
\toprule
 & \multicolumn{4}{c}{\textbf{Memory content}} & \multicolumn{3}{c}{\textbf{Construction conditions}} & \multicolumn{3}{c}{\textbf{Query time}} \\
\cmidrule(lr){2-5}\cmidrule(lr){6-8}\cmidrule(lr){9-11}
\textbf{Method} & \shortstack{Handled\\objects} & \shortstack{Untouched\\objects} & \shortstack{Motion\\history} & \shortstack{Path \&\\places}
 & \shortstack{RGB\\only} & \shortstack{Scene\\changes} & \shortstack{$\geq$20-min\\video}
 & \shortstack{No frames\\at query} & \shortstack{Language\\queries} & \shortstack{3D\\location} \\
\midrule
ConceptGraphs~\citep{gu2024conceptgraphs}          & \xmark & \cmark & \xmark & \xmark & \xmark & \xmark & N.A.   & \cmark & \cmark & \cmark \\
3D-Mem~\citep{yang2025_3dmem}                      & \xmark & \cmark & \xmark & \xmark & \xmark & \xmark & N.A.   & \xmark & \cmark & \cmark \\
ReMEmbR~\citep{anwar2024remembr}                   & \xmark & \xmark & \xmark & \cmark & \xmark & N.A.   & \cmark & \cmark & \cmark & \pmark \\
VideoAgent~\citep{fan2024videoagent}               & \pmark & \cmark & \xmark & \xmark & \cmark & \xmark & \xmark & \xmark & \cmark & \xmark \\
AMEGO~\citep{goletto2024amego}                     & \cmark & \xmark & \pmark & \cmark & \cmark & \xmark & \cmark & \cmark & \xmark & \xmark \\
OSNOM~\citep{Plizzari2025OSNOM}                    & \cmark & \xmark & \cmark & \xmark & \pmark & \xmark & \cmark & N.A.   & \xmark & \cmark \\
Emb.\ VideoAgent~\citep{fan2025embodied}          & \cmark & \cmark & \pmark & \xmark & \xmark & \xmark & \xmark & \xmark & \cmark & \cmark \\
EgoRAG~\citep{yang2025egolife}                     & \pmark & \xmark & \xmark & \xmark & \cmark & \cmark & \cmark & \xmark & \cmark & \xmark \\
VL-KnG~\citep{almdfaa2025vlkng}                    & \xmark & \cmark & \pmark & \cmark & \cmark & \xmark & \cmark & \pmark & \cmark & \xmark \\
DirectMe~\citep{wang2026ucsbench}                  & \pmark & \cmark & \pmark & \cmark & \cmark & \xmark & \cmark & \xmark & \cmark & \cmark \\
R4DSG$^{a}$~\citep{ma2026r4dsg}                   & \pmark & \pmark & \cmark & \pmark & \cmark & \pmark & \cmark & \cmark & \cmark & \xmark \\
\midrule
\rowcolor{gray!12}\textbf{\sysname (ours)}         & \cmark & \cmark & \cmark & \cmark & \cmark & \cmark & \cmark & \cmark & \cmark & \cmark \\
\bottomrule
\end{tabular}
\\[2pt]\begin{minipage}{0.98\textwidth}\scriptsize
$^{a}$R4DSG's objects enter through a per-segment controlled vocabulary and its places through location segments, both taken from released segment records whose source the paper does not state; anchors are found per location window, and the authors report limited cross-day persistence.
\end{minipage}
\end{table}

%% file: sections/related_work.tex
\section{Related Work}
\label{sec:related}
\begin{figure}[t]
\centering
\includegraphics[width=\textwidth]{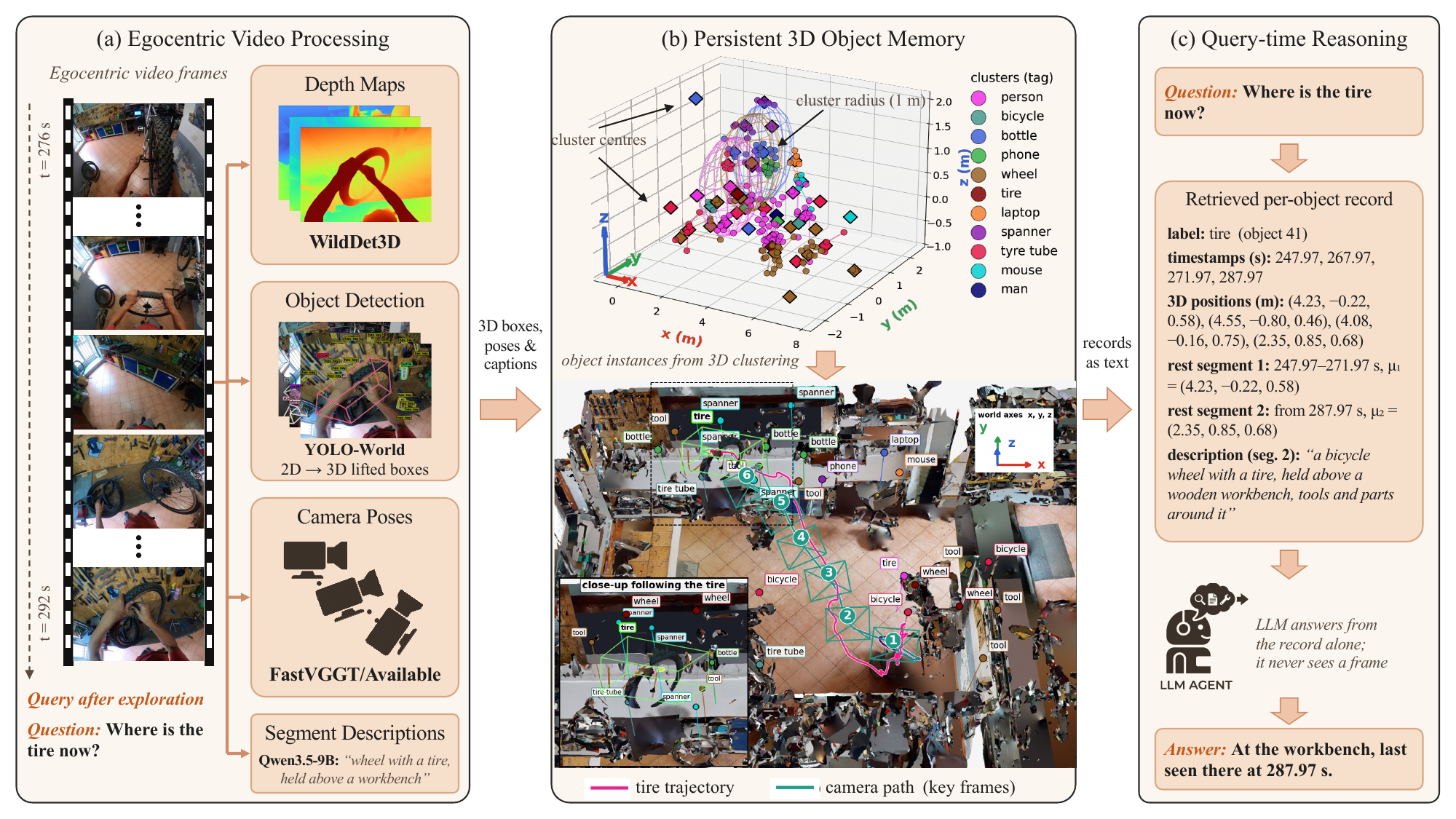}
\caption{\sysname overview. (a) Objects detected in sampled egocentric frames
are localized in 3D and placed in a scene coordinate frame using supplied or
estimated camera poses. (b) Persistent object
records, formed from associated observations, contain timestamped positions, rest segments defined by repeated
evidence of displacement, and contextual descriptions. (c) After query grounding
and retrieval, a language model answers from the records rendered as text,
without receiving images. When pose scale is available, scene coordinates are metric.}
\label{fig:method}
\end{figure}

\paragraph{Memories of objects and scenes.}
OSNOM tracks the 3D locations of active objects through absence with appearance
and spatial association \citep{Plizzari2025OSNOM}. Whareformer learns association
with an updatable track memory and an explicit new-track decision and is trained on
active objects. However, its architecture does not require contact
\citep{chalk2026whareformer}. AMEGO builds location
memories and interaction tracklets for querying on long videos \citep{goletto2024amego}, and ESOM
couples discovery, tracking, and retrieval in an online visual-query memory
\citep{manigrasso2024esom}. These systems determine which objects enter memory and how to recognize
them when they reappear. Scene representations also differ in the information
they store. ConceptGraphs builds open-vocabulary scene graphs of objects
from posed RGB-D observations \citep{gu2024conceptgraphs}. Snapshot and
scene-based memories support spatial reasoning and exploration
\citep{yang2025_3dmem,lu2026gsmem,zhou2026memory}. Khronos records short-term
motion and long-term scene changes \citep{schmid2024khronos}. DirectMe tracks
objects, places, and the wearer, and uses geometry to derive spatial relations
for timestamped questions \citep{wang2026ucsbench}.

We study how the objects retained, their estimated positions, movement
histories, and contextual descriptions help answer different spatial questions.
Errors in geometry can misplace an object or make it appear to have moved.
We therefore evaluate how these errors affect answers, alongside their
effect on localization.

\paragraph{Reading stored memories.}
Memory-based systems differ in what they store, how they find relevant
information, and whether they show images to the answerer. ReMEmbR stores
captions with timestamps and positions, then searches this memory over
multiple rounds to answer a question \citep{anwar2024remembr}. VideoAgent
and Embodied VideoAgent combine stored information about entities with
retrieved frames \citep{fan2024videoagent,fan2025embodied}. MEMORA maintains
entity state for later reasoning \citep{yu2026memora}, while DirectMe gives
the answerer its graph and retrieved frames \citep{wang2026ucsbench}.
Video-language and streaming models use sampled frames or compressed visual
tokens without explicit object records
\citep{cheng2024videollama2,zhang2024longva,zhang2024llavavideo,geminiteam2024gemini15,zhang2025flashvstream,qian2025dispider,huang2025videochatonline}.
We test what stored records can support when the answerer receives only text.
We also compare our object records with event descriptions against a caption
memory using the same retrieval procedures. This helps distinguish the value
of the stored information from the reader's ability to find and use it.

\paragraph{Evaluating spatial memory over time.}
Existing benchmarks test different demands on memory. \egofour \vqthreed
queries an object's location using an image crop \citep{grauman2022ego4d};
\egoloc, ReLoCATE, and EAGLE address this task through pose recovery,
training-free retrieval, and appearance and geometric memory
\citep{mai2023egoloc,khosla2025relocate,cao2025eagle}. We evaluate localization
using object names instead. \hdepic tests object and fixture positions and
motion on calibrated recordings \citep{perrett2025hdepic}, while \ucs tests
timestamped spatial reasoning relative to the moving wearer
\citep{wang2026ucsbench}. EASG-Bench tests questions grounded in scene graphs \citep{rodin2025easg},
while OpenEQA tests questions about previously observed environments
\citep{majumdar2024openeqa}. HourVideo evaluates reasoning over long egocentric
videos \citep{chandrasegaran2024hourvideo}. EgoMemReason extends this to
evidence spanning days, including changes in object states
\citep{wang2026egomemreason}. We complement these benchmarks by asking the
same questions on individual recordings and on streams formed by joining
recordings. We then modify memory construction and retrieval to identify
why performance changes.

%% file: sections/method.tex
\section{Method}
\label{sec:method}

\sysname (Long-horizon Egocentric Descriptions, Geometry, and Event Records) constructs a persistent object memory offline from an egocentric
recording and supplied or estimated camera poses (Figure~\ref{fig:method}), once per
recording and independently of any question. It is then reused for spatial question
answering and named-object localization; on \ucs{}, reading it costs about 2.0k
input tokens per question, against 9.6k for 48 sampled frames
(Appendix~\ref{app:cost}).

For each instance $j$, we store its label $c_j$ and timestamped observations
$\mathcal{O}_j=\{(t_i,b_i,\mathbf{x}_i)\}$, where $b_i$ is the image box and
$\mathbf{x}_i$ is the estimated 3D position in scene coordinates. We summarize the object's observed location history
as \emph{rest segments},
$\mathcal{S}_j=\{(t_s^{\rm start},t_s^{\rm end},\boldsymbol\mu_s,d_s)\}$.
Each segment groups observations associated with one resting location:
$t_s^{\rm start}$ and $t_s^{\rm end}$ delimit the segment's span in the
sampled observation sequence, $\boldsymbol\mu_s$ represents its estimated
3D location, and $d_s$ describes the object and its surrounding context.
A new segment is confirmed only after repeated observations support a
displacement from the current location. Rest segments thus summarize
where the object was observed; its position during gaps between
observations is unknown. Camera poses also retain the
wearer's trajectory and heading for egocentric queries. Unrelated scenes are not
registered into a shared frame; coordinates are metric where pose scale is
available.

\subsection{Offline memory construction}
\label{sec:method-offline}

\textbf{Discovery and geometry.} \tagger names objects in sampled frames;
synonym merging produces a vocabulary for \detectorfull, so objects enter memory
whether or not the wearer touches them, within the detection and 3D-lifting budgets.
From the image, detection box, and label, \lifterfull predicts a 3D box in camera
coordinates, and the camera pose places its centre in scene coordinates
(Appendix~\ref{app:construction}).

\textbf{Association and persistence.} We match observations in time order to
existing instances by label and distance from the instance's latest position;
same-frame detections stay separate unless they are overlapping duplicates. A second
pass re-identifies objects that 3D association split into several tracks:
\trackerfull propagates each fragment's masks and merges fragments whose masks agree,
subject to co-occurrence and 3D-distance checks
(Appendix~\ref{app:construction-association}). Records persist after objects leave
view, although a large move may create a new identity. A new rest segment starts
only after $k$ successive observations lie more than $\tau$ from the current resting
position; observations that return within this radius stay in the current segment.
This suppresses false moves from localization noise but may miss brief ones.

\textbf{Context and refinement.} \tagger describes selected segments from frames with
the object outlined, recording contents, supporting surfaces, and interactions (on
\hdepic, transferred from distance-threshold segments;
Appendix~\ref{app:construction-segments}). Multi-view triangulation refines a resting
object's position when the viewpoints are well separated and the estimate is stable
and close to the original (Appendix~\ref{app:construction-refinement}). Two
extensions add activity descriptions for sampled frames and build a separate memory
for each detected scene (Appendices~\ref{app:ucs-ablations} and~\ref{app:stitch}).

\subsection{Query-time reasoning}
\label{sec:method-online}

\textbf{Grounding and retrieval.} \hdepic box queries are lifted to a 3D point on
the sampled frame nearest their timestamp, and the four nearest instances observed
within $\pm6$\,s are retrieved; named fixtures are matched through fixed synonym
groups, taking the most-observed matches, with clock directions computed from stored
positions and the camera heading at query time. \vqthreed supplies an object name,
and we return the selected instance's stored position; \ucs targets are inferred
from the question, and entries are ranked by word overlap and temporal proximity
(Appendices~\ref{app:implementation}, \ref{app:vq3d}, and~\ref{app:ucs}).

\textbf{Answering.} Retrieved labels, ranking signals, and segment histories (times,
positions, and descriptions) are rendered as text, and a single call to
\qwen~\citep{qwen35blog} or \gpt~\citep{openai2026gpt54thinking} answers from this
evidence, the question, and its options. \hdepic queries use the completed memory;
for timestamped \ucs queries, snapshots remove later observations and descriptions,
truncate histories and the wearer path, and recompute positions. This filters what
the answerer sees without making construction causal
(Appendix~\ref{app:ucs-causality}).

%% file: sections/experiments.tex
\section{Experiments}
\label{sec:evaluation}

We first show three memories answering on one stream; five research questions
then connect benchmark results to interventions on the memory, and
Appendix~\ref{app:loss} analyses failures by stage.

\paragraph{Benchmarks and baselines.} We use all 2,400 \catperception and
\catmotion questions of \hdepic~\citep{perrett2025hdepic} with prototype-averaged
accuracy (Appendix~\ref{app:hdepic-prototypes}); all 164 \egofour \vqthreed
validation queries~\citep{grauman2022ego4d}, localized from the object's name and
scored over returned predictions alongside coverage; and 2,771 \ucs
questions~\citep{wang2026ucsbench} from 211 EgoLife and TeleEgo recordings, with
estimated poses and memories filtered to the question time
(Appendix~\ref{app:ucs-protocol}). Evidence conditions are a blind control
(question and options only), memory alone, 48 frames sampled up to the question
time, and frames plus memory on spatial questions. DirectMe, ReMEmbR, OSNOM, and
AMEGO are rebuilt from released code with dataset adaptations and read by \gpt
unless stated; RQ4 examines their retrieval.

\begin{figure}[tbp]
\centering
\includegraphics[width=0.9\textwidth]{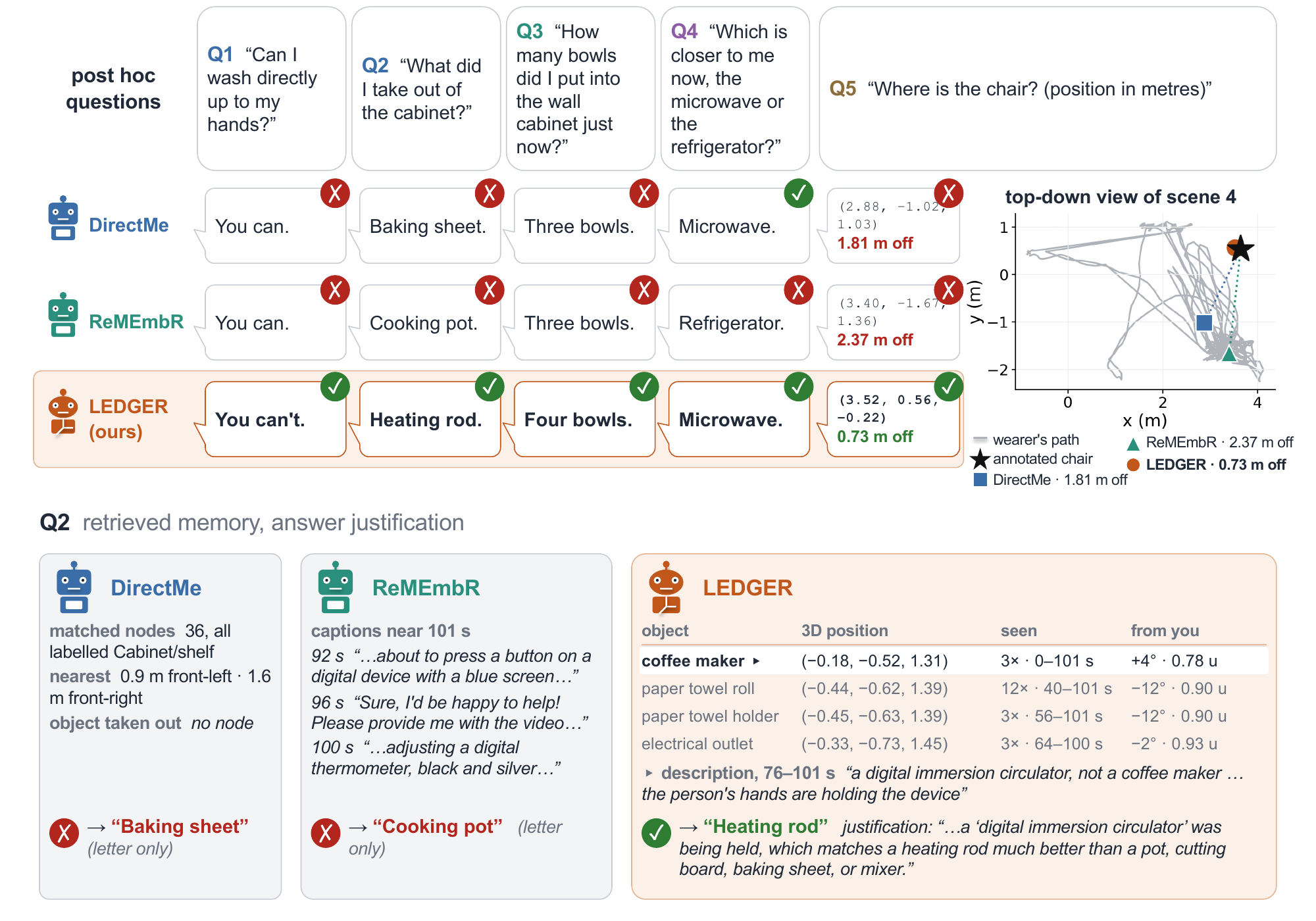}
\caption{Three memories answer the same questions from their stored records alone,
without frames. Q1--Q4 come from one \ucs stream that joins three EPIC-Kitchens
recordings; Q5 asks for a chair's position in an \egofour \vqthreed clip, with each
memory's answer on the top-down map. \textbf{Bottom:} what each memory retrieves for
Q2. Questions were chosen to show how the memories differ;
Figure~\ref{fig:qualitative} (Appendix~\ref{app:qualitative}) adds the video timeline.}
\label{fig:qual-main}
\end{figure}

\paragraph{Three memories on one stream.} Figure~\ref{fig:qual-main} follows DirectMe, ReMEmbR, and \sysname through one
stream. \sysname is correct on all five questions, DirectMe's graph on one, and
ReMEmbR on none, and the difference follows from what each memory keeps. For Q2,
the detector's label is wrong, but the description stored with the object's rest
segment names what the wearer was holding, which the answerer follows; neither
baseline stored the object. For Q5, \sysname returns
the chair's position, 0.73\,m from the annotation, whereas ReMEmbR returns only
where the wearer stood.

\begin{table}[tbp]
\centering
\footnotesize
\setlength{\tabcolsep}{3pt}
\caption{\hdepic accuracy (\%) on all 2,400 \catperception and \catmotion
questions, prototype-averaged as in \citet{perrett2025hdepic}; chance is 20.0.
\vidyes: video at answer time; \vidno: text alone. Published rows use other
answerers and protocols, and the 2026 entry is a challenge report
(Appendix~\ref{app:hdepic-baselines}). \sysnameagent: our memory read by ReMEmbR's
querying agent (\ref{sec:sensitivity}). Bold: best per evidence group, except the
human reference; 95\% intervals in Table~\ref{tab:hdepic-ci}; other answerers in
Figure~\ref{fig:components}b.}
\label{tab:hdepic-main}
\begin{tabular}{lcccc}
\toprule
Method & Video at answer & \catperception & \catmotion & Mean \\
\midrule
LongVA~\citep{zhang2024longva}                                     & \vidyes & 32.9 & 22.7 & 27.8 \\
2025 challenge, 1st place~\citep{taluzzi2025pixels}                & \vidyes & 42.6 & 30.2 & 36.4 \\
EgoAdapt, 2026~\citep{chen2026egoadapt}                            & \vidyes & \textbf{64.9} & \textbf{61.6} & \textbf{63.3} \\
Human (sample)~\citep{perrett2025hdepic}                           & \vidyes & 93.8 & 92.7 & 93.3 \\
\midrule
\multicolumn{5}{l}{\emph{Memories rebuilt on our frames, read by \gpt}} \\
DirectMe, graph only~\citep{wang2026ucsbench}                      & \vidno  & 30.6 & 30.8 & 30.7 \\
ReMEmbR~\citep{anwar2024remembr}                                   & \vidno  & 28.7 & 30.9 & 29.8 \\
OSNOM, adapted~\citep{Plizzari2025OSNOM}                           & \vidno  & 35.4 & 25.0 & 30.2 \\
AMEGO, adapted~\citep{goletto2024amego}                            & \vidno  & 30.4 & 38.9 & 34.7 \\
\midrule
\gpt, blind                                                        & \vidno  & 27.2 & 32.1 & 29.7 \\
\sysname + \gpt                                                    & \vidno  & \textbf{44.7} & \textbf{40.5} & \textbf{42.6} \\
\sysnameagent + \gpt                                      & \vidno  & 37.0 & 32.6 & 34.8 \\
\bottomrule
\end{tabular}
\end{table}

\begin{figure}[tbp]
\centering
\IfFileExists{plots/fig_components.pdf}{\includegraphics[width=\textwidth]{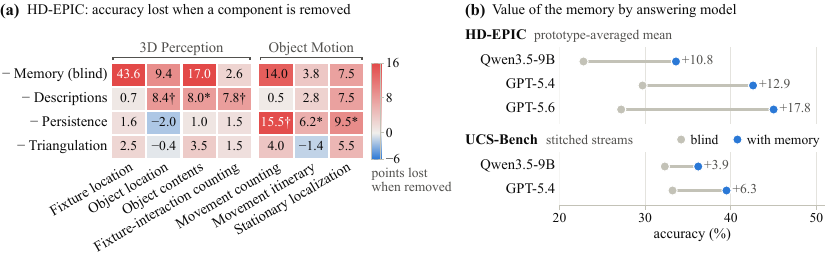}}{\fbox{\parbox{0.9\textwidth}{\centering\vspace{5em}\todo{plots/fig\_components.pdf pending}\vspace{5em}}}}
\caption{What each component carries, and what the answering model adds.
(a)~\hdepic: accuracy lost on each prototype when one component is removed
($^{\dagger}$ survives Holm correction, $^{*}$ raw $p<0.05$; single runs, cells
$\pm$7--10 points). (b)~Accuracy without and with the memory for each answering
model; GPT-5.6 answered with reasoning enabled, the others without. The \ucs
component and reader analyses are in Figure~\ref{fig:components-ucs}.}
\label{fig:components}
\end{figure}

\begin{table}[tbp]
\centering
\footnotesize
\setlength{\tabcolsep}{2.5pt}
\caption{\ucs accuracy (\%) on 2,771 timestamped questions; 48 frames where
frames are given; chance 25.7. DM is DirectMe
with its released graph builder and retrieval: ``own'' adds its graph to the frames
on every question, as DirectMe does, and ``routed'' only on the spatial questions
where we add our memory. Bold: best per row among the green and among the red arms.
Subcategories and paired tests are in Appendix~\ref{app:ucs-protocol}.}
\label{tab:ucs-main}
\begin{tabular}{lrcc cc ccc}
\toprule
 & & & & \multicolumn{2}{c}{\novidsq\textbf{Memory only}} & \multicolumn{3}{c}{\vidsq\textbf{Frames + memory}} \\
\cmidrule(lr){5-6}\cmidrule(lr){7-9}
Category & $n$ & \novidsq Blind & \vidsq Frames & DM & \sysname & DM, own & DM, routed & \sysname \\
\midrule
Trajectory \& Movement    & 648  & 34.9 & 45.4 & 37.3 & \textbf{38.6} & 41.8 & 45.5 & \textbf{49.1} \\
Position \& Orientation   & 701  & 23.8 & 39.7 & 26.2 & \textbf{33.4} & 37.4 & 37.4 & \textbf{43.9} \\
Proximity \& Reachability & 538  & 45.4 & \textbf{51.7} & 40.0 & \textbf{49.3} & 47.2 & 48.5 & 51.3 \\
Category \& Quantity      & 884  & 33.9 & 42.9 & 21.4 & \textbf{36.0} & 35.6 & 42.0 & \textbf{43.1} \\
\midrule
Overall                   & 2771 & 33.8 & 44.4 & 30.0 & \textbf{38.5} & 39.8 & 42.9 & \textbf{46.3} \\
\bottomrule
\end{tabular}
\end{table}

\paragraph{RQ1: What can object memory answer, and what evidence does it miss?}\rqlabel{RQ1}{sec:hdepic-results}\label{sec:hdepic-protocol}
\emph{Object records help every answering model we tested, most on objects the
wearer never touched and least on interactions and counts.} On \hdepic, the memory
lifts \qwen, \gpt, and GPT-5.6 by 11 to 18 points over their blind controls and,
read by \gpt, outscores every rebuilt memory (Table~\ref{tab:hdepic-main};
Figure~\ref{fig:components}b); on \ucs, it also improves the 48-frame reader it is
added to (Table~\ref{tab:ucs-main}). The clearest benefit is fixture location,
because admitting untouched objects supplies evidence that contact-gated memories
exclude. The weakest cases need events and identities rather than more objects:
AMEGO's hand--object tracklets remain better on interaction counting and the onset
of rest (Appendix~\ref{app:hdepic-membase}), and \ucs object counts do not improve
because association both splits single objects into several tracks and misses others
(Appendix~\ref{app:ucs-counting}).

\paragraph{RQ2: When does localization noise get confused for movement?}\rqlabel{RQ2}{sec:hdepic-ablation}\label{sec:evidence}
\emph{Requiring repeated evidence of a move, rather than raising the distance
threshold, keeps localization noise from being counted as movement.} A threshold
alone cannot separate the two, because they overlap in size: half of the annotated
moves are shorter than 0.8\,m, and the 95th percentile of localization error is
0.84\,m. A 0.3\,m threshold therefore roughly doubles the counted moves, and 0.8\,m
gets only 57.3\% of counts within $\pm1$ of the truth. Keeping 0.3\,m but requiring
three consecutive displaced observations raises this to 67.4\%
(Appendix~\ref{app:hdepic-motion}), and removing persistence costs movement
counting 15.5 points, which survives correction for multiple tests
(Figure~\ref{fig:components}a). The price is that brief moves go unrecorded.

\begin{table}[tbp]
\centering
\footnotesize
\setlength{\tabcolsep}{3pt}
\caption{\novidsq \vqthreed validation split: memories built by each method on the
same 44 clips and read by the same \gpt matcher from the object's name; errors in
metres over graded queries (Appendix~\ref{app:vq3d-harness}). AMEGO stores neither
names nor 3D positions. ReMEmbR answers with the wearer's position;
OSNOM uses our detections and depth; DirectMe is read without frames. Our rows name
the detector and use trajectory memory and \egoloc poses, without \segmenter
consolidation.}
\label{tab:vq3d-harness}
\begin{tabular}{lrrrrr}
\toprule
Memory & Graded / 164 & Median $L_2$ (m) & Mean $L_2$ (m) & $<1$\,m (\%) & $<2$\,m (\%) \\
\midrule
ReMEmbR (wearer position)               & 164 & 1.49 & 1.77 & 23 & 69 \\
DirectMe (graph only)                   & 164 & 1.91 & 1.98 & 18 & 52 \\
OSNOM (adapted)                         & 146 & 1.33 & 1.65 & 35 & 71 \\
\emph{No memory: clip centroid}                      & 164 & 1.69 & 1.82 & 10 & 63 \\
\midrule
\sysname, \detector                                  & 148 & 1.06 & 1.46 & 49 & 74 \\
\sysname, \segmenter                                 & 153 & \textbf{0.99} & \textbf{1.41} & \textbf{51} & \textbf{76} \\
\sysnameagent, \segmenter                          & 164 & 1.00 & 1.54 & 49 & \textbf{76} \\
\bottomrule
\end{tabular}
\end{table}

\paragraph{RQ3: Which geometric errors matter for spatial answering?}\rqlabel{RQ3}{sec:baselines}\label{sec:vq3d}
\emph{Errors in the camera reference frame matter more for answering than errors
in object coordinates.} Direction answers depend on the camera's heading, so
estimated poses cost fixture-location questions 11 points, or 29 without gravity
levelling (Appendix~\ref{app:hdepic-design}). Refining object coordinates by
triangulation, by contrast, adds only 2.2 points overall, less than persistence (5.5)
or descriptions (4.9; Figure~\ref{fig:components}a and Table~\ref{tab:hdepic-ablation}). Coordinates are still what make
metric localization possible: on \vqthreed, our memory has the lowest error of the
rebuilt memories on its returned predictions (Table~\ref{tab:vq3d-harness}). Even
there, pose quality bounds the refinement: triangulation improves eligible
predictions by about 0.1\,m with scan-registered poses but only 0.03\,m with
feed-forward poses (Appendix~\ref{app:vq3d-triangulation}).

\paragraph{RQ4: How much does retrieval affect comparisons between memories?}\rqlabel{RQ4}{sec:sensitivity}
\emph{Which memory looks better depends on the reader, so comparisons must hold
the reader fixed.} We read our record with two readers: our single-shot reader (one
call, about 2k input tokens) and ReMEmbR's querying agent (\sysnameagent; three to
four calls of 16--23k tokens; Appendix~\ref{app:ucs-reader}). When the same agent
reads both memories, our record beats ReMEmbR's captions by 5.0 points on \hdepic
($p<10^{-8}$) and by 26 points in \vqthreed queries within 1\,m, and is level on
\ucs (Tables~\ref{tab:hdepic-main}, \ref{tab:vq3d-harness}, and~\ref{tab:ucs-reader}). ReMEmbR leads
natively only on \ucs (41.4 against 39.0), and the lead disappears in two steps:
event lines on what the wearer did, the weakness found in RQ1, bring our record to
41.0 with our reader (Table~\ref{tab:stitch}), and the shared agent brings it to 42.1
(Appendix~\ref{app:ucs-reader}), as on stitched streams
(Figure~\ref{fig:components-ucs}b). The agent is not the better reader everywhere: on
\hdepic, where our reader grounds box-marked questions in 3D, it beats the agent on the
same record by 7.8 points over all 2,400 questions (Table~\ref{tab:hdepic-main}) at a
thirteenth of the tokens.

\begin{figure}[tbp]
\centering
\includegraphics[width=0.85\textwidth]{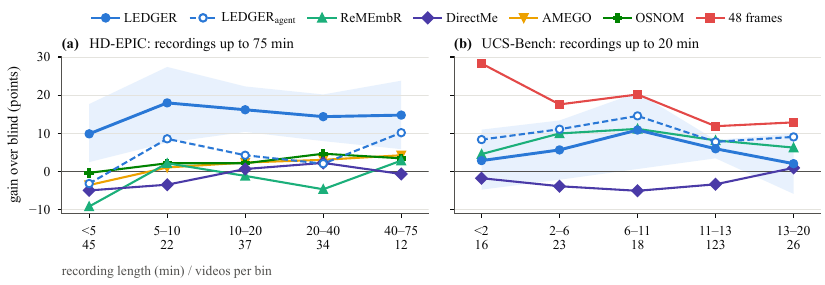}
\caption{Gain over the blind control by recording length, on the same questions in
each bin (question-level accuracy, \gpt). The band is the 95\% interval of \sysname's
gain. (a)~\hdepic, 2,400 questions on
150 videos. (b)~\ucs, 3,264 questions on 206 single recordings; AMEGO and OSNOM
cannot answer these. Per-bin tables and trend tests are in Appendix~\ref{app:length}.}
\label{fig:length}
\end{figure}

\paragraph{RQ5: What breaks as recordings grow longer or span different scenes?}\rqlabel{RQ5}{sec:retention}
\emph{Longer recordings do not erode the memory's advantage, but changes of scene
do.} Over length, the memory stays 10 to 18 points above the blind control in every
\hdepic bin up to 75 minutes, while no rebuilt memory gains more than 5; on \ucs, the
48-frame reader loses accuracy as recordings grow and the memory does not
(Figure~\ref{fig:length}; Appendix~\ref{app:length}). Across scenes, 100 paired
stitched streams show that returning to the same scene costs the memory little,
whereas a change of scene brings it to the blind level (Table~\ref{tab:stitch}).
The loss arises in both stages, because each assumes one scene: retrieval shows
records and histories from other scenes, and construction drops labels that cannot
belong to one place while its pose gate rejects the whole stream at the seam. Fixing
either stage recovers part of the loss. Limiting retrieved records and histories to
the current scene recovers 3.0 points with true boundaries, and building the memory
per detected scene recovers about half in emulation (Appendices~\ref{app:stitch-construction} and~\ref{app:stitch}). The fixes
are not additive, and these streams also change the camera, wearer, and clip length.

\FloatBarrier %
\section{Conclusion and Limitations}
\label{sec:conclusion}

\sysname makes past object observations reusable for spatial answering and
localization. Admitting untouched objects answers fixture questions, and persistence
and contextual descriptions contribute more than coordinate refinement; missing event
evidence and fragmented identities limit interaction and counting. Comparisons
between memories depend on the reader, and scene changes break both retrieval and
construction, which per-scene construction partly repairs. Evaluations should
measure answerable evidence alongside localization, and should state retrieval budgets and
scene boundaries.

The scope remains limited. Recent frame readers outperform memory-only answering on
\hdepic, and frames alone outperform memory alone on \ucs, although adding the memory
to the frames improves them. Counting and continuous rest intervals remain
unresolved, and construction does not yet scale to the most object-dense hour-long
recordings (Appendix~\ref{app:ucs}). \hdepic uses calibrated poses; \vqthreed is
queried by name rather than crop, and most \ucs memories lack metric scale.
On \hdepic, the segmentation and retrieval rules, triangulation thresholds, and fixture
synonym groups were developed on calibration videos kept separate from evaluation
(Appendix~\ref{app:hdepic-dev}).
Construction is offline, so filtering
records by question time does not establish causal online memory.

%% file: sections/appendix_index.tex
\addtocontents{toc}{\protect\setcounter{tocdepth}{2}}
\renewcommand{\contentsname}{Appendix Contents}
\begingroup
\small
\setlength{\parskip}{0pt}
\makeatletter
\renewcommand*\l@section[2]{%
  \ifnum \c@tocdepth >\z@
    \addpenalty\@secpenalty
    \addvspace{0.5em \@plus\p@}%
    \setlength\@tempdima{1.5em}%
    \begingroup
      \parindent \z@ \rightskip \@pnumwidth
      \parfillskip -\@pnumwidth
      \leavevmode \bfseries
      \advance\leftskip\@tempdima
      \hskip -\leftskip
      #1\nobreak\hfil
      \nobreak\hb@xt@\@pnumwidth{\hss #2\kern-\p@\kern\p@}\par
    \endgroup
  \fi}
\makeatother
\tableofcontents
\endgroup

\subsection*{Reading guide}
\begin{center}
\small
\begin{tabular}{@{}p{0.56\textwidth}p{0.40\textwidth}@{}}
\toprule
Main text & Appendices with the details \\
\midrule
Method (Section~\ref{sec:method}) & \ref{app:construction} \\
Benchmarks, baselines, and protocols (Section~\ref{sec:evaluation})
 & \ref{app:hdepic-prototypes}--\ref{app:hdepic-membase}, \ref{app:vq3d-protocol}--\ref{app:vq3d-implementation},
   \ref{app:vq3d-published}--\ref{app:vq3d-harness}, \ref{app:ucs-protocol}, \ref{app:ucs-implementation},
   \ref{app:stitch-design} \\
RQ1: which questions stored records answer (Section~\ref{sec:evaluation})
 & \ref{app:hdepic-prototypes}, \ref{app:hdepic-baselines}--\ref{app:hdepic-membase},
   \ref{app:ucs-protocol}--\ref{app:ucs-counting}, \ref{app:ucs-answerers} \\
RQ2: genuine movement or observation noise (Section~\ref{sec:evaluation})
 & \ref{app:hdepic-ablation}, \ref{app:hdepic-motion} \\
RQ3: which geometric errors matter (Section~\ref{sec:evaluation})
 & \ref{app:hdepic-ablation}--\ref{app:hdepic-geometry}, \ref{app:vq3d} \\
RQ4: retrieval and memory comparisons (Section~\ref{sec:evaluation})
 & \ref{app:ucs-ablations}, \ref{app:ucs-reader} \\
RQ5: what breaks across scenes (Section~\ref{sec:evaluation})
 & \ref{app:length}, \ref{app:stitch-design}--\ref{app:stitch}, \ref{app:delayed} \\
Where wrong answers are lost; a worked example (Section~\ref{sec:evaluation})
 & \ref{app:loss}, \ref{app:qualitative} \\
Why a memory rather than frames per question (Section~\ref{sec:method})
 & \ref{app:cost} \\
\bottomrule
\end{tabular}
\end{center}
\clearpage

%% file: sections/appendix_method.tex
\section{Memory Construction Details}
\label{app:construction}

\subsection{Sampling, vocabulary, and detection}
We sample one frame every four seconds, up to 1,200 frames per recording.
\tagger names the visible objects. We merge synonymous names into a
vocabulary of at most 80 labels per recording. On \ucs{}, this step uses
at most 80 frames (Appendix~\ref{app:ucs}). \detectorfull then detects
objects matching these labels. We also evaluate \segmenterfull as an
alternative detector on \vqthreed{}. Which objects enter memory depends
on the vocabulary, the detector's confidence threshold, and the number
of detections we can lift into 3D. Dataset-specific resolutions,
thresholds, and budgets are listed in Appendices~\ref{app:implementation},
\ref{app:vq3d}, and~\ref{app:ucs}.

\subsection{Camera poses and object lifting}
We use calibrated device trajectories on \hdepic{} and registered \egoloc{}
poses on \vqthreed{}. To estimate poses, we use \geometryfull{}, a
training-free acceleration of \geometrybase{}. It processes 96-frame
chunks with 16 overlapping frames. We align adjacent chunks using a
Sim(3) transformation estimated from their shared frames.

These estimated poses have arbitrary scale. To recover metric scale,
the pipeline compares pairs of observations of the same static object
with their predicted metric positions in camera coordinates. This step
fails for most \ucs{} recordings and all \teleego{} recordings.
Distances in those memories are therefore expressed in scene units
(Appendix~\ref{app:ucs-protocol}).

On \ucs{}, we first try \depthanything~\citep{lin2025depthanything3},
except for circular-fisheye recordings. It processes 48-frame chunks
with 12 overlapping frames. We keep its poses only if they pass three
consistency checks: the chunk-alignment residual is at most 0.5\,m,
chunk scales lie within $[0.8,1.25]$, and the camera moves no more than
6\,m between sampled frames. Otherwise, we fall back to \geometry{}.
If \geometry{} cannot align a chunk with the preceding one, it starts
from a new anchor (Appendix~\ref{app:ucs}).

Given an image, a 2D box, and an object label, \lifterfull predicts a
metric 3D box in camera coordinates. We use its centre as the object's
position. This avoids selecting pixels from a dense depth map, where
object and background surfaces can be mixed. Although the predicted box
is metric, its position in scene coordinates is only metric if the
camera poses also have a known scale.

\paragraph{Geometry-estimator analysis.}
In the earlier \hdepic{} analysis, the median localization error was
0.39\,m, with a median along-ray component of 0.37\,m
(Appendix~\ref{app:hdepic-geometry}). This motivates using multiple
views with known camera poses to refine object positions.

On the \ucs{} pilot, two \depthanything{} reconstructions were rejected.
Their chunk scale ratios were 1.16 and 1.58, and their alignment residuals
at chunk boundaries were 0.67 and 7.10\,m, respectively. We fall back to
\geometry{}, which also handles fisheye recordings but does not provide
metric scale on its own.

We compare both pose estimators on the same frames from all 44
\vqthreed{} clips (Appendix~\ref{app:vq3d}). After aligning each trajectory
to \egoloc{} camera centres with one Sim(3) transformation per clip,
their median position errors are similar: 9.1\,cm for \depthanything{}
and 9.2\,cm for \geometry{}. Median orientation errors are
5.5$^{\circ}$ and 5.9$^{\circ}$, respectively. However, \depthanything{}
fails to join its chunks into a single trajectory on 11 clips.
Object localization also shows no clear difference: median paired
differences range from $-0.03$ to $+0.02$\,m across detectors and memory
variants, with every interval including zero. These results support
using \geometry{} as the fallback when poses are unavailable, although
both estimators still require metric scale recovery.

\subsection{Association and consolidation}
\label{app:construction-association}
Trajectory memory clusters observations in 3D, associating them chronologically
by label and proximity to an instance's most recent position. Detections in the same frame remain separate unless they are overlapping
duplicates. A second pass, called consolidation, reconnects tracks that
may belong to the same object. It groups tracks with synonymous labels,
uses \trackerfull{} to propagate their masks across frames, and merges
tracks whose propagated masks agree. Co-occurrence and 3D-distance checks
block incompatible merges.

We use consolidation on \hdepic{} and \ucs{}. On \vqthreed{}, we compare
raw, trajectory, and consolidated memories, using a trajectory association
radius of 1.0\,m and a consolidation mask-IoU threshold of 0.3.
On \ucs{}, merges are blocked beyond a 3D distance of 1.0, measured in
scene units when metric scale is unavailable
(Appendix~\ref{app:ucs-counting}). Even with consolidation, an object
may receive a new identity after a long absence or a large move.

\subsection{Persistence and contextual descriptions}
\label{app:construction-segments}
Distance-threshold segmentation starts a new segment as soon as an
observation lies more than $\tau$ from the current resting position.
Our persistence rule waits until $k$ successive observations of the
object lie outside this radius before confirming a new segment.
If observations return within the radius before confirmation, they
remain in the current segment.

On \hdepic{}, we use $\tau=0.3$\,m and $k=3$. We use the same values
on \ucs{}, but measure the radius in scene units when metric scale
is unavailable. Appendix~\ref{app:hdepic-motion} compares the two rules.

For up to six segments per object, \tagger{} describes a representative
frame with the object outlined. These descriptions record details such
as the object's contents, its supporting surface, and interactions
that positions alone do not capture.

In the reported \hdepic{} runs, we first generated descriptions for
distance-threshold segments. Each persistence segment then inherited
the description of the segment with the greatest temporal overlap.
The persistence ablation therefore also changes which descriptions
appear in the records; it does not keep all textual evidence fixed
(Appendix~\ref{app:implementation}). The additional activity descriptions
for sampled frames in the \ucs{} extensions are detailed in
Appendix~\ref{app:ucs-ablations}.

\subsection{Multi-view refinement}
\label{app:construction-refinement}
Observing a resting object from several viewpoints helps constrain its
3D position. Each detected box centre defines a viewing ray from the
camera. Given camera centres $\mathbf{c}_i$ and unit ray directions
$\mathbf{d}_i$, we estimate the point that minimizes the sum of squared
distances to these rays:
\begin{equation}
  \hat{\mathbf{x}}=
  \left(\sum_i\mathbf{P}_i\right)^{-1}\sum_i\mathbf{P}_i\mathbf{c}_i,
  \qquad \mathbf{P}_i=\mathbf{I}-\mathbf{d}_i\mathbf{d}_i^{\top}.
  \label{eq:triangulation}
\end{equation}
We accept this estimate only when the camera positions are sufficiently
separated, the ray geometry supports a stable solution, and the estimate
stays close to the monocular prediction. On \hdepic{}, we require at
least three rays, a camera baseline of at least 0.10\,m, a minimum
eigenvalue of 0.25 for $\sum_i\mathbf{P}_i$, and a shift of no more than
2.0\,m from the monocular prediction (Table~\ref{tab:hdepic-hparams}).
The localization analysis and question-answering ablation are reported
in Appendices~\ref{app:hdepic-geometry} and~\ref{app:hdepic-ablation}.

%% file: sections/appendix_hd_epic.tex
\section{\hdepic Details}
\label{app:hdepic}

\subsection{Evaluation scope and per-prototype results}
\label{app:hdepic-subset}
\label{app:hdepic-prototypes}

\begin{figure}[htbp]
\centering
\begin{minipage}[t]{0.49\textwidth}\centering
\includegraphics[width=\linewidth]{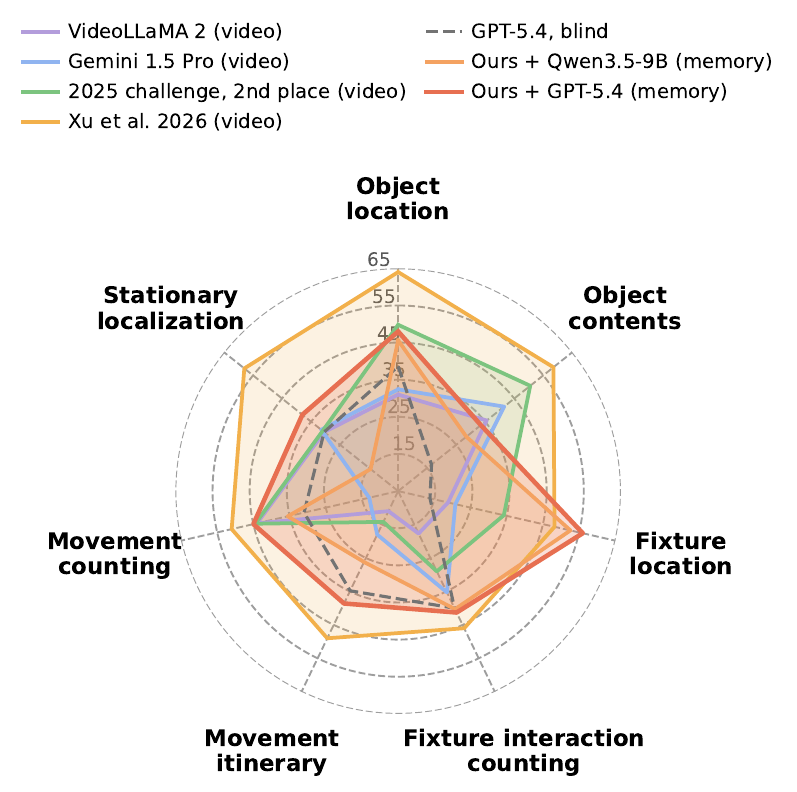}
\end{minipage}\hfill
\begin{minipage}[t]{0.49\textwidth}\centering
\includegraphics[width=\linewidth]{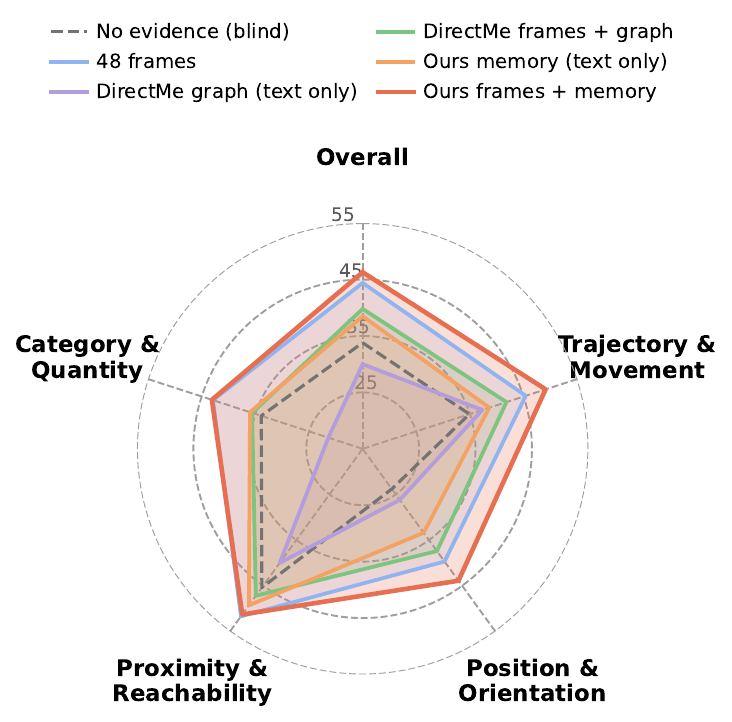}
\end{minipage}
\caption{Accuracy by question type. \textbf{Left:} \hdepic by prototype; published
video models as printed in \citet{xu2026semantic} (EgoAdapt reports no per-prototype
scores). \textbf{Right:} \ucs by category on the 2,771 questions of Table~\ref{tab:ucs-main}.}
\label{fig:radar}
\end{figure}

\hdepic{} contains 26,550 five-way multiple-choice questions across
30 prototypes and seven categories. We evaluate all questions in
\catperception{} and \catmotion{}, covering 150 videos
(Table~\ref{tab:hdepic-prototype}; Figure~\ref{fig:radar}).
These questions have no official training, validation, or test split.

We build memories using the released camera poses and reserve ground-truth
geometry for analysis. Across the 150 videos (40.7 hours), 36,633 sampled
frames produce 52,001 object instances, 504,981 observations, and 82,445
rest segments. Answering from the same memory in a separate session
yields 41.9\% accuracy, compared with 42.6\% originally, illustrating
a 0.7-point variation between these two runs.

Component ablations use 2,317 questions from the 133 videos processed
in the initial pass, with the same prototype-averaged metric.
The earlier ten-video analyses reported below are development studies
based on different run snapshots.

\begin{table}[htbp]
\centering
\small
\setlength{\tabcolsep}{4pt}
\caption{\hdepic{} accuracy (\%) on all seven evaluated prototypes.
Memory scores are bold when they exceed the matched blind control.}
\label{tab:hdepic-prototype}
\begin{tabular}{llrcccc}
\toprule
& & & \multicolumn{2}{c}{\qwen} & \multicolumn{2}{c}{\gpt} \\
Category & Prototype & $n$ & Blind & Memory & Blind & Memory \\
\midrule
\multirow{4}{*}{\catperception}
 & object location & 500 & 25.4 & \textbf{45.6} & 38.6 & \textbf{48.0} \\
 & contents retrieval & 200 & 16.0 & \textbf{28.5} & 16.5 & \textbf{33.5} \\
 & fixture location & 500 & 16.2 & \textbf{52.6} & 13.8 & \textbf{56.0} \\
 & fixture interaction counting & 300 & 36.0 & \textbf{40.3} & 40.0 & \textbf{41.3} \\
\midrule
\multirow{3}{*}{\catmotion}
 & movement itinerary & 500 & 11.8 & \textbf{26.2} & 34.8 & \textbf{38.6} \\
 & movement counting & 200 & 28.5 & \textbf{35.5} & 31.0 & \textbf{45.0} \\
 & stationary localization & 200 & 26.0 & 14.5 & 30.5 & \textbf{38.0} \\
\midrule
\multicolumn{2}{l}{Overall (question average)} & 2,400 & 21.5 & \textbf{37.5} & 29.7 & \textbf{44.6} \\
\bottomrule
\end{tabular}
\end{table}

\begin{table}[htbp]
\centering
\small
\caption{Prototype-averaged \hdepic{} accuracy (\%) with 95\% bootstrap intervals
(2,000 resamples stratified by prototype, seed 0). Memory improves mean
accuracy over the matched blind control by 10.8 points $[8.2,13.3]$
for \qwen{} and 13.0 points $[10.4,15.5]$ for \gpt{}.}
\label{tab:hdepic-ci}
\begin{tabular}{llccc}
\toprule
Answerer & Evidence & \catperception & \catmotion & Mean \\
\midrule
\multirow{2}{*}{\qwen}
 & Blind & 23.4 {\scriptsize[21.2,25.7]} & 22.1 {\scriptsize[19.0,25.1]} & 22.8 {\scriptsize[20.9,24.7]} \\
 & Memory & \textbf{41.8} {\scriptsize[39.3,44.4]} & \textbf{25.4} {\scriptsize[22.3,28.4]} & \textbf{33.6} {\scriptsize[31.6,35.5]} \\
\multirow{2}{*}{\gpt}
 & Blind & 27.2 {\scriptsize[25.0,29.6]} & 32.1 {\scriptsize[28.9,35.4]} & 29.7 {\scriptsize[27.7,31.7]} \\
 & Memory & \textbf{44.7} {\scriptsize[42.0,47.3]} & \textbf{40.5} {\scriptsize[37.2,44.3]} & \textbf{42.6} {\scriptsize[40.5,44.8]} \\
\bottomrule
\end{tabular}
\end{table}

We report 95\% bootstrap intervals using 2,000 resamples stratified by
prototype (seed 0). Comparisons use the same questions in both conditions.

We also report pooled accuracy, which gives each question equal weight,
so larger prototypes contribute more. For \qwenshort{}, memory raises
pooled \catperception{} accuracy from 23.2\% to 44.6\% and \catmotion{}
accuracy from 18.7\% to 25.7\%. For \gptshort{}, the corresponding
improvements are from 27.7\% to 47.4\% and from 33.0\% to 39.9\%.

Across all 2,400 questions, pooled accuracy rises from 21.5\% to 37.5\%
for \qwenshort{} and from 29.7\% to 44.6\% for \gptshort{}. The paired
gains are 16.0 points $[13.6,18.3]$ and 14.9 points $[12.7,17.3]$,
respectively. These pooled scores should not be compared directly with
the \hdepic{} paper, which averages prototype scores.

The intervals describe uncertainty for the evaluated videos. They do
not account for clustering of questions within videos or replace
evaluation on held-out videos.

\subsection{Implementation and query adapters}
\label{app:implementation}

We convert the Aria fisheye RGB frames into $1802\times1802$ virtual
pinhole images. We combine the released device poses with the camera
calibration to obtain the world-to-camera transformation:
$\mathbf{T}_{cw}=(\mathbf{T}_{wd}\mathbf{T}_{dc})^{-1}$.
To check the coordinate conventions, we project 35,722 ground-truth
object locations into the images. Of these, 98.3\% fall inside their
annotated boxes, with a median offset of 22 pixels. These annotations
are not used to build the memory.

\begin{table}[htbp]
\centering
\small
\setlength{\tabcolsep}{4pt}
\caption{\hdepic construction and answering settings.}
\label{tab:hdepic-hparams}
\begin{tabular}{lll}
\toprule
Stage & Parameter & Value \\
\midrule
Sampling & interval / cap & 4\,s / 1,200 frames \\
Rectification & pinhole resolution & $1802\times1802$ px \\
Tagging & model / token cap / vocabulary cap & \tagger{} / 130 / 80 \\
Detection & model & \detector v8x-worldv2 \\
 & input / confidence / top-$k$ & 1408 px / 0.05 / 40 \\
3D lifting & per-frame / per-video / per-tag budget & 12 / $\geq400$ / $\max(25,N/4)$ \\
Triangulation & rays / baseline / $\lambda_{\min}$ / shift & 3 / 0.10\,m / 0.25 / 2.0\,m \\
Association & tracker IoU / 3D merge veto & 0.3 / 1.0\,m \\
Segmentation & radius $\tau$ / persistence $k$ & 0.3\,m / 3 \\
Descriptions & frame size / segments per object & 896 px / 6 \\
Box retrieval & time window / candidates & $\pm6$\,s / 4 \\
Answering & protocol & one call, options A--E \\
\bottomrule
\end{tabular}
\end{table}

For questions that specify an object with a box, we use the sampled frame
nearest the query time to estimate the box's 3D position. We rank object
instances observed near that time by their minimum distance to this position
and retrieve the closest four. We then replace the box reference in the
question with the estimated coordinate. The answerer receives object labels,
ranking signals, and segment histories as text. It sees no images, although
grounding the query box still uses a frame.

For fixture-location questions, we match names using predefined synonym
groups, such as hob/stove/cooktop and fridge/refrigerator/freezer.
We retrieve the four matching instances with the most observations.
To compute their clock directions, we project the camera's forward axis
and the direction to each object onto the horizontal plane: 12 is ahead,
3 is right, 6 is behind, and 9 is left.

Using ground-truth fixture positions, this calculation selects the correct
direction option on 97.8\% of 365 validation questions. With the constructed
memory, retrieval finds a name-matched fixture for 83\% of the 500
fixture-location questions. Choosing the option closest to the first
candidate's clock direction scores 59.2\% across all 500 questions,
compared with 56.0\% for \gptshort{} reading the candidate histories.
This geometric rule applies specifically to fixture-direction questions.

We run \qwenshort{} with greedy decoding and thinking disabled.
We run \gpt{} with reasoning effort set to none and temperature 1.0
in all experiments, so its answers can vary across otherwise identical
runs. Mean usage per question is 1,938 input and 381 output tokens for
\qwenshort{}, and 1,750 input and 231 output tokens for \gptshort{}.

Descriptions are generated for distance-threshold segments. Each
persistence segment inherits the description of the segment with the
greatest temporal overlap. The segmentation ablation therefore changes
both segment boundaries and the assignment of descriptions; it does
not keep all textual evidence fixed.

\subsection{Published baselines and aggregation}
\label{app:hdepic-baselines}

The \hdepic{} paper averages prototype accuracies to obtain category
scores \citep{perrett2025hdepic}. We follow this convention for our
primary results. We also report pooled accuracy, which gives each
question equal weight. For category $C$, with prototype accuracies
$a_p$ and question counts $n_p$, the two scores are
\[
\text{Prototype average}=\frac{1}{|C|}\sum_{p\in C}a_p,
\qquad
\text{Pooled accuracy}=
\frac{\sum_{p\in C}n_p a_p}{\sum_{p\in C}n_p}.
\]
Our rows in Table~\ref{tab:hdepic-baselines} use prototype averages
to match the published convention. The number of correct answers
for each prototype can be uniquely recovered from its rounded
accuracy and question count in Table~\ref{tab:hdepic-prototype}.

\begin{table}[htbp]
\centering
\footnotesize
\setlength{\tabcolsep}{2.5pt}
\caption{Published \hdepic category scores and our prototype-averaged scores
(\%). Published baselines were not rerun; models, visual inputs, and inference
budgets differ. The team labels
used in the challenge reports are inconsistent between papers, so rows are
identified by paper: the 40.9 / 29.9 entry is the method of \citet{yang2026hdepic}
(2nd place 2025), and the 42.6 / 30.2 entry is the scene- and knowledge-graph
system of \citet{taluzzi2025pixels} (1st place 2025, 44.21 overall).}
\label{tab:hdepic-baselines}
\begin{tabular}{lccc}
\toprule
Model & Images to answerer & \catperception & \catmotion \\
\midrule
Llama 3.2, text only~\citep{perrett2025hdepic}            & \vidno  & 22.3 & 25.5 \\
Gemini 1.5 Pro, text only~\citep{perrett2025hdepic}       & \vidno  & 21.5 & 27.7 \\
VideoLLaMA 2~\citep{cheng2024videollama2}                 & \vidyes & 25.7 & 28.5 \\
LongVA~\citep{zhang2024longva}                            & \vidyes & 32.9 & 22.7 \\
LLaVA-Video~\citep{zhang2024llavavideo}                   & \vidyes & 27.3 & 18.9 \\
Gemini 1.5 Pro~\citep{geminiteam2024gemini15}             & \vidyes & 32.5 & 20.8 \\
Qwen2.5-VL-7B, zero-shot~\citep{yang2026hdepic}           & \vidyes & 35.0 & 23.9 \\
Qwen2.5-VL-32B, zero-shot~\citep{yang2026hdepic}          & \vidyes & 35.6 & 19.8 \\
2025 challenge, 2nd place~\citep{yang2026hdepic}          & \vidyes & 40.9 & 29.9 \\
2025 challenge, 1st place~\citep{taluzzi2025pixels}       & \vidyes & 42.6 & 30.2 \\
Gemini 3.1 Pro + object detector, 2026~\citep{xu2026semantic} & \vidyes & 54.2 & 52.7 \\
EgoAdapt, 2026~\citep{chen2026egoadapt}                   & \vidyes & 64.9 & 61.6 \\
Human (sample)~\citep{perrett2025hdepic}                  & \vidyes & 93.8 & 92.7 \\
\midrule
\qwen blind (ours) & \vidno & 23.4 & 22.1 \\
\sysname + \qwen & \vidno & 41.8 & 25.4 \\
\gpt blind (ours) & \vidno & 27.2 & 32.1 \\
\sysname + \gpt & \vidno & 44.7 & 40.5 \\
\bottomrule
\end{tabular}
\end{table}

\begin{table}[htbp]
\centering
\footnotesize
\setlength{\tabcolsep}{3pt}
\caption{\hdepic accuracy per prototype (\%) for the published entries that
report it (as printed in \citet{xu2026semantic}) and for our memory. Bold marks
the best score per prototype. EgoAdapt reports category means only.}
\label{tab:hdepic-proto-baselines}
\begin{tabular}{lrcccccccc}
\toprule
Prototype & $n$ & \shortstack{Video-\\LLaMA 2} & \shortstack{LLaVA-\\Video} & \shortstack{Gemini\\1.5 Pro} & \shortstack{2025\\2nd} & \shortstack{Xu et al.\\2026} & \shortstack{\gptshort\\blind} & \shortstack{Ours +\\\qwenshort} & \shortstack{Ours +\\\gptshort} \\
\midrule
Fixture interaction counting & 300 & 17.7 & 16.3 & 35.3 & 29.0 & \textbf{46.0} & 40.0 & 40.3 & 41.3 \\
Fixture location             & 500 & 18.8 & 21.8 & 20.8 & 34.2 & 48.2 & 13.8 & 52.6 & \textbf{56.0} \\
Object location              & 500 & 31.0 & 30.6 & 32.4 & 49.8 & \textbf{64.0} & 38.6 & 45.6 & 48.0 \\
Object contents retrieval    & 200 & 35.5 & 40.5 & 41.5 & 50.5 & \textbf{58.5} & 16.5 & 28.5 & 33.5 \\
Movement itinerary           & 500 & 11.0 &  9.8 & 18.0 & 14.2 & \textbf{49.0} & 34.8 & 26.2 & 38.6 \\
Movement counting            & 200 & 44.0 & 20.0 & 13.0 & 44.5 & \textbf{51.0} & 31.0 & 35.5 & 45.0 \\
Stationary object localization & 200 & 30.5 & 27.0 & 31.5 & 31.0 & \textbf{58.0} & 30.5 & 14.5 & 38.0 \\
\bottomrule
\end{tabular}
\end{table}

Using \gptshort, the memory exceeds the video baselines of the \hdepic paper and
both 2025 challenge entries under this averaging convention, and it remains below
the two 2026 challenge reports, which read the video with newer models. EgoAdapt
uses category-specific video sampling and option-order voting. Neither comparison isolates the benefit of memory; our matched blind
runs provide that control. We also tested a newer model on ten
development videos covering 418 questions, using the same answering
protocol with three evidence conditions: 50 frames, memory, or no
evidence. Prototype-averaged \catperception{} and \catmotion{} scores
were 57.9\% and 45.8\% with frames, 40.1\% and 33.4\% with memory,
and 23.6\% and 32.4\% for the blind control. Frames outperform memory
in this test, consistent with the published 2026 results.

The per-prototype results reveal where memory helps most
(Table~\ref{tab:hdepic-proto-baselines}). Fixture location is the only
prototype where our memory exceeds every reported baseline score.
Movement counting and fixture-interaction counting are comparable
to the 2025 entries. The weakest results are on object contents
retrieval and, for \qwenshort{}, stationary object localization.

\subsection{Memories rebuilt on our setup}
\label{app:hdepic-membase}

Four memory-based methods are rebuilt on the same 150 videos and 2,400 questions.
Each method uses its released code to build memory from the same rectified
frames, sampled every 4\,s. \gpt{} answers from each memory without seeing
the video, and all scores use the same prototype averaging. For questions
that identify an object with a box, we match the query box to each memory's
own observations at the query time. We select up to three matches with
the highest 2D overlap, requiring IoU $\geq0.1$. Method-specific adaptations:
\begin{itemize}
\item \textbf{DirectMe} \citep{wang2026ucsbench}: released graph builder (Objects365
vocabulary, \detector, SAM~2, \depthanything depth and poses) and its text-only
graph prompt without keyframes, evaluated over the whole video, with the pose for
egocentric relations taken at the question's time. Its object-count shortcut is
disabled because \hdepic counting questions count events, not instances.
\item \textbf{ReMEmbR} \citep{anwar2024remembr}: released VILA captioner (one caption
per sampled frame), camera positions from the released device trajectories, and its
retrieval agent with the options appended; memory times are video times.
\item \textbf{OSNOM} \citep{Plizzari2025OSNOM}: released tracker with its published
configuration, given the detections our memory lifts (as box masks), the released
device trajectories, and \geometry depth scaled to metres per 120-frame chunk in
place of its mesh-rendered depth. It has no question-answering stage; a common
harness serializes the matched track's position history, nearby tracks, and, for
fixture questions, name-matched tracks with clock directions computed as in our agent.
\item \textbf{AMEGO} \citep{goletto2024amego}: its hand--object detector weights are no
longer available, so the Hands23 detector~\citep{cheng2023hands23} from the same group replaces
it. Its frame-count thresholds, set for 50--60\,fps, give an empty memory at
0.25\,fps and are reduced to one frame; flow filtering and the object tracker are
disabled. AMEGO stores no object names and no 3D positions.
\end{itemize}
The 95\% bootstrap intervals on the mean are $[28.8,32.8]$ for DirectMe,
$[27.8,31.8]$ for ReMEmbR, $[28.2,32.2]$ for OSNOM, and $[32.6,36.7]$ for AMEGO,
against $[40.5,44.8]$ for our memory.

\begin{table}[htbp]
\centering
\small
\setlength{\tabcolsep}{4pt}
\caption{\hdepic accuracy per prototype (\%, rounded) for the memories rebuilt on our
setup. Bold marks the best memory per prototype.}
\label{tab:hdepic-membase-proto}
\begin{tabular}{lrcccccc}
\toprule
Prototype & $n$ & Blind & DirectMe & ReMEmbR & OSNOM & AMEGO & Ours \\
\midrule
Fixture location               & 500 & 14 & 26 & 21 & 44 & 13 & \textbf{56} \\
Object location                & 500 & 39 & 37 & 39 & 42 & 40 & \textbf{48} \\
Object contents retrieval      & 200 & 16 & 20 & 30 & 21 & 22 & \textbf{34} \\
Fixture interaction counting   & 300 & 40 & 39 & 25 & 35 & \textbf{47} & 41 \\
Movement itinerary             & 500 & 35 & 16 & 14 & 19 & 27 & \textbf{39} \\
Movement counting              & 200 & 31 & 37 & 32 & 16 & 34 & \textbf{45} \\
Stationary object localization & 200 & 30 & 39 & 47 & 40 & \textbf{56} & 38 \\
\bottomrule
\end{tabular}
\end{table}

\subsection{Component ablations}
\label{app:hdepic-ablation}

``Segment descriptions'' removes all descriptions and summaries from the final
memory; ``persistence segmentation'' uses the distance-threshold rule
($\tau=0.3$\,m, no persistence); ``multi-view triangulation'' uses monocular depth
with the same persistence segmentation. Table~\ref{tab:hdepic-ablation-ci} reports category-level effects and
their intervals. Per-prototype comparisons use the same questions
in both conditions (Table~\ref{tab:hdepic-ablation}). We use exact
McNemar tests, with Holm correction across the 21 prototype--component
comparisons, and bootstrap intervals from 4,000 resamples of videos.

Three effects remain significant after correction: descriptions improve
object location by 8.4 points $[3.0,13.4]$ and fixture-interaction
counting by 7.8 points $[3.8,12.0]$; persistence improves movement
counting by 15.5 points $[8.1,23.2]$.

Each condition uses a single \gpt{} run. Repeating identical prompts
changes whether 15--20\% of answers are correct, and individual
per-prototype scores vary by about $\pm7$--10 points.

\begin{table}[htbp]
\centering
\footnotesize
\setlength{\tabcolsep}{4pt}
\caption{Accuracy lost on each \hdepic prototype when one component is removed from
the memory (full $-$ ablated, points; 2,317 questions from 133 videos).
$^{\dagger}$Significant after Holm correction over the 21 tests; $^{*}$raw
$p<0.05$. One run per arm, so single cells vary by about $\pm$7--10 points.}
\label{tab:hdepic-ablation}
\begin{tabular}{lr rr rrr}
\toprule
Prototype & $n$ & Full & Blind & $-$Descriptions & $-$Persistence & $-$Triangulation \\
\midrule
\multicolumn{7}{l}{\emph{\catperception}} \\
Fixture location                 & 447 & 56.4 & 12.8 & $+0.7$ & $+1.6$ & $+2.5$ \\
Object location                  & 500 & 48.0 & 38.6 & $+8.4^{\dagger}$ & $-2.0$ & $-0.4$ \\
Object contents retrieval        & 200 & 33.5 & 16.5 & $+8.0^{*}$ & $+1.0$ & $+3.5$ \\
Fixture interaction counting     & 270 & 42.2 & 39.6 & $+7.8^{\dagger}$ & $+1.5$ & $+1.5$ \\
\multicolumn{7}{l}{\emph{\catmotion}} \\
Movement counting                & 200 & 45.0 & 31.0 & $+0.5$ & $+15.5^{\dagger}$ & $+4.0$ \\
Movement itinerary               & 500 & 38.6 & 34.8 & $+2.8$ & $+6.2^{*}$ & $-1.4$ \\
Stationary localization          & 200 & 38.0 & 30.5 & $+7.5$ & $+9.5^{*}$ & $+5.5$ \\
\midrule
\catperception (prototype mean)  & 1,417 & 45.0 & 26.9 & $+6.2$ & $+0.5$ & $+1.8$ \\
\catmotion (prototype mean)      & 900   & 40.5 & 32.1 & $+3.6$ & $+10.4$ & $+2.7$ \\
Mean                             & 2,317 & 42.8 & 29.5 & $+4.9$ & $+5.5$ & $+2.2$ \\
\bottomrule
\end{tabular}
\end{table}

\begin{table}[htbp]
\centering
\small
\setlength{\tabcolsep}{5pt}
\caption{\hdepic component ablations at the category level (questions and runs of
Table~\ref{tab:hdepic-ablation}; prototype-averaged mean of the two categories).
Differences are paired, computed before rounding, with 95\% stratified bootstrap
intervals.}
\label{tab:hdepic-ablation-ci}
\begin{tabular}{lccc}
\toprule
Component removed & Without & With & Gain (points) \\
\midrule
Persistence segmentation & 37.3 & 42.8 & $+5.5$ {\scriptsize[+3.1,+7.9]} \\
Segment descriptions & 37.9 & 42.8 & $+4.9$ {\scriptsize[+2.9,+6.9]} \\
Multi-view triangulation & 40.5 & 42.8 & $+2.2$ {\scriptsize[+0.2,+4.4]} \\
\bottomrule
\end{tabular}
\end{table}

\subsection{Design ablations}
\label{app:hdepic-design}

Each row of Table~\ref{tab:hdepic-design} changes one part of the final \hdepic
system and re-answers all 2,400 questions with the same prompts. We test differences in mean accuracy using paired sign-flip permutation
tests with 20,000 draws and Holm correction across the twelve design
comparisons. For individual prototypes, we use exact McNemar tests
with Holm correction across 84 cells. Only the estimated poses without
gravity levelling significantly change the mean after correction.
The GPT-5.6, text-retrieval, and 2D-area-retrieval differences are
significant only before correction.

The pose-free variants use \geometry{} poses. We rescale their camera
centres by $1/1.70$, a factor measured against ground-truth depth on
\hdepic{}. These variants therefore avoid the released camera poses
but still use ground-truth information for scale. Gravity levelling
estimates the up direction as the direction most nearly perpendicular
to all camera $x$-axes, with a median error of 6.0$^{\circ}$.

The GPT-5.6 blind control includes six API errors and seven unparsed
answers, all scored as incorrect. These affect its score by at most
0.3 points. GPT-5.6 uses reasoning on 98.9\% of questions, with a
median of 366 reasoning tokens. In contrast, \gpt{} uses no reasoning
tokens. The 2.4-point gain therefore combines a change of model with
the addition of reasoning.

The bottom block reports full memory rebuilds on a ten-video pilot
covering 418 questions. Each is compared with the final memory
re-answered in the same session. Intervals are approximately
$\pm5$ points, and no difference is significant. Sampling every
two seconds mainly lowers movement-counting accuracy, but this
needs further checking before drawing a conclusion. Across the
981 annotated object names in these videos, semantic recall is
61\% for \tagger{}, 60\% for \gpt{} with vision, and 54\% for
Gemma-3-4B.

The component removals in Table~\ref{tab:hdepic-ablation} come from
a separate experiment on 2,317 questions with its own full-memory
baseline. They are not repeated here.

\begin{table}[htbp]
\centering
\scriptsize
\setlength{\tabcolsep}{2pt}
\caption{\hdepic design ablations: one change from the final system at a time,
all 2,400 questions, prototype-averaged accuracy. $\Delta$ is against the final
system; the prototype columns give the change on each prototype.
Markers and single-run noise as in Table~\ref{tab:hdepic-ablation}, with Holm
correction over this table's tests. $^{\S}$See the text for the pose rescaling.}
\label{tab:hdepic-design}
\begin{tabular}{l rr rrrr rrr}
\toprule
& & & \multicolumn{4}{c}{\catperception} & \multicolumn{3}{c}{\catmotion} \\
\cmidrule(lr){4-7}\cmidrule(lr){8-10}
Change from the final system & Mean & $\Delta$ & obj.\ loc. & contents & fixt.\ loc. & fixt.\ count & itinerary & mov.\ count & stationary \\
\midrule
Final system (\gpt) & 42.6 & -- & 48.0 & 33.5 & 56.0 & 41.3 & 38.6 & 45.0 & 38.0 \\
\midrule
Answering model: GPT-5.6 & 45.0 & $+2.4^{*}$ & $+0.4$ & $+7.0^{*}$ & $+2.2$ & $+0.7$ & $+5.2^{*}$ & $+0.5$ & $+1.0$ \\
\quad no memory, \gpt & 29.7 & $-13.0^{*}$ & $-9.4^{*}$ & $-17.0^{*}$ & $-42.2^{*}$ & $-1.3$ & $-3.8$ & $-14.0^{*}$ & $-7.5$ \\
\quad no memory, GPT-5.6 & 27.2 & $-15.4^{*}$ & $-20.0^{*}$ & $-14.5^{*}$ & $-36.8^{*}$ & $-1.0$ & $-21.0^{*}$ & $-25.5^{*}$ & $+8.5$ \\
\midrule
Lift depth: \geometry & 44.7 & $+2.1$ & $+2.2$ & $-5.0$ & $+5.2^{*}$ & $-1.7$ & $+7.8^{*}$ & $+1.5$ & $+2.5$ \\
\quad \lifter, depth $\div$ 1.70 & 43.0 & $+0.4$ & $+1.8$ & $-3.5$ & $+2.6$ & $+0.3$ & $+9.0^{\dagger}$ & $-1.5$ & $-6.0$ \\
\quad \depthanything & 42.0 & $-0.7$ & $+1.0$ & $-4.5$ & $+1.6$ & $-1.0$ & $+6.2^{*}$ & $-0.5$ & $-7.5$ \\
\midrule
Poses: \geometry, gravity-levelled$^{\S}$ & 40.8 & $-1.9$ & $+4.4^{*}$ & $+0.5$ & $-11.4^{\dagger}$ & $-2.3$ & $+1.0$ & $+2.0$ & $-7.5$ \\
\quad same, not levelled$^{\S}$ & 39.1 & $-3.5^{\dagger}$ & $+3.8^{*}$ & $+2.0$ & $-29.4^{\dagger}$ & $-2.0$ & $+1.0$ & $+3.0$ & $-6.0$ \\
\midrule
Retrieval: 2D box IoU & 42.0 & $-0.6$ & $+2.6$ & $+1.0$ & $+0.4$ & $-2.7$ & $+0.8$ & $-3.0$ & $-2.5$ \\
\quad 3D box IoU & 41.0 & $-1.6$ & $-0.6$ & $-0.5$ & $-1.2$ & $+0.0$ & $+1.2$ & $-2.0$ & $-7.0$ \\
\quad 2D area within 1\,m & 40.4 & $-2.2^{*}$ & $-1.8$ & $+0.5$ & $-0.6$ & $+0.0$ & $+0.6$ & $-5.5$ & $-7.0$ \\
\quad text only & 39.6 & $-3.0^{*}$ & $+1.8$ & $-8.5^{*}$ & $+0.0$ & $-2.3$ & $-5.2$ & $+0.5$ & $-6.5$ \\
Question rephrasing off & 41.6 & $-1.1$ & $+0.8$ & $-1.0$ & $+0.8$ & $-3.3$ & $-0.8$ & $-1.0$ & $-2.5$ \\
\midrule
\multicolumn{10}{l}{\emph{Whole-memory rebuilds on a ten-video pilot (418 questions; reference: final system re-answered, 33.4)}} \\
Pilot reference (final system) & 33.4 & -- & 53.2 & 16.7 & 34.0 & 34.4 & 39.3 & 33.3 & 24.3 \\
Detector: \segmenter & 36.3 & $+2.9$ & $-3.2$ & $+12.5$ & $+12.8$ & $+3.1$ & $+1.2$ & $-2.7$ & $+0.0$ \\
Tagger: Gemma-3-4B & 32.8 & $-0.6$ & $-5.1$ & $+4.1$ & $+0.0$ & $-3.2$ & $+4.7$ & $+0.0$ & $-5.4$ \\
\quad \gpt vision & 34.9 & $+1.5$ & $-6.4$ & $+0.0$ & $+0.0$ & $+6.2$ & $+3.6$ & $+0.0$ & $+5.4$ \\
Sampling: one frame per 2\,s & 28.8 & $-4.7$ & $-5.1$ & $-4.2$ & $+2.2$ & $+0.0$ & $-6.0$ & $-16.6$ & $+0.0$ \\
\bottomrule
\end{tabular}
\end{table}

\subsection{Earlier retrieval, oracle, and geometry analyses}
\label{app:hdepic-geometry}

The following analyses use an earlier 10-video development snapshot with 339
object-level questions. These results come from separate runs, rather than the current
2,400-question evaluation or the 2,317-question component ablations.
Unless stated otherwise, the configurations in
Table~\ref{tab:hdepic-retrieval} use distance-threshold segmentation
and monocular positions. With \gptshort{}, retrieval criteria differ
by at most 0.3 accuracy points. Switching the answerer changes
accuracy by 7.4 points when using the same 3D-centre retrieval.

\begin{table}[htbp]
\centering
\small
\caption{Earlier 339-question retrieval analysis (accuracy, \%).}
\label{tab:hdepic-retrieval}
\begin{tabular}{lcc}
\toprule
Retrieval & \qwen & \gpt \\
\midrule
Blind control & 19.5 & -- \\
3D centre distance & 29.2 & 36.6 \\
2D area among candidates within 1\,m & 33.9 & 36.6 \\
Union of both sets & 30.4 & 36.3 \\
3D centre, triangulated positions & -- & 40.4 \\
\bottomrule
\end{tabular}
\end{table}

\begin{table}[htbp]
\centering
\small
\caption{Earlier 339-question oracle analysis. Ground-truth
supporting-fixture labels add information beyond correcting coordinates.}
\label{tab:hdepic-oracle}
\begin{tabular}{lc}
\toprule
Evidence & Accuracy (\%) \\
\midrule
Predicted memory, triangulated positions & 40.4 \\
Ground-truth queried-object trajectory, positions only & 52.2 \\
Ground-truth trajectory and supporting-fixture names & 77.6 \\
Any of four single-candidate answers, retrieved objects & 65.5 \\
Any of four single-candidate answers, non-retrieved objects & 64.9 \\
\bottomrule
\end{tabular}
\end{table}

The trajectory oracle (Table~\ref{tab:hdepic-oracle}) corrects target selection and history as well as localization.
It does not separately identify each failure source. The near-identical retrieved
and non-retrieved ``any of four'' scores also show why best-of-several answering
success cannot be interpreted as retrieved evidence coverage: multiple guesses can
produce a high score without selecting relevant objects.

We evaluate triangulation on 389 detections matched to ground-truth
boxes with IoU $\geq0.5$. It reduces median 3D error from 0.39 to
0.35\,m and increases the share within 0.25\,m of the true position
from 13.6\% to 37.3\% (Equation~\ref{eq:triangulation}). Before
refinement, most error lies along the viewing ray: the median
along-ray component is 0.37\,m, compared with 0.03\,m perpendicular
to it.

For 377 detections observed from at least two known camera poses,
we group results into quartiles by camera baseline. The share
within 0.25\,m improves in every group: from 10.5\% to 27.4\%,
12.8\% to 46.8\%, 12.7\% to 39.2\%, and 19.1\% to 38.3\%.
This analysis measures localization of matched detections, not
how many objects the system discovers.

\subsection{Movement and stationary-state analysis}
\label{app:hdepic-motion}

\begin{table}[htbp]
\centering
\small
\caption{Earlier movement-count analysis on 89 questions whose queried object
is in memory (ground-truth mean count: 2.5).}
\label{tab:hdepic-segmentation}
\begin{tabular}{lcccc}
\toprule
Rule & Mean count & Exact & Within $\pm1$ & Bias \\
\midrule
Threshold, $\tau=0.3$\,m & 4.9 & 13.5\% & 36.0\% & +2.3 \\
Threshold, $\tau=0.8$\,m & 2.3 & 22.5\% & 57.3\% & $-0.3$ \\
Persistence, $\tau=0.3$\,m, $k=3$ & 2.2 & \textbf{28.1\%} & \textbf{67.4\%} & $-0.3$ \\
\bottomrule
\end{tabular}
\end{table}

The threshold rule over-counts movement in 64\% of these questions at 0.3\,m
(Table~\ref{tab:hdepic-segmentation}).
A larger distance threshold suppresses localization noise but can miss
small movements. In the annotations, 19\% of moves are shorter than
0.3\,m and 50\% are shorter than 0.8\,m. Meanwhile, the 95th percentile
of localization error is 0.84\,m. Persistence improves movement counts
on questions whose target is already in memory, separating segmentation
errors from failures to retain the target. Bias and mean counts are
rounded separately.

Stationary-localization questions ask when an object stopped moving
and remained still. In the earlier analysis, choosing the answer
timestamp closest to a rest-segment start is correct for only 22\%
of questions, compared with 20\% chance. This suggests that sampled
observation times do not reliably identify when continuous rest began.
Memory does not always hurt this question type, however: the current
\qwenshort{} and \gptshort{} runs show opposite effects
(Table~\ref{tab:hdepic-prototype}).

\subsection{Development data and cost}
\label{app:hdepic-dev}

We developed the segmentation and retrieval rules, triangulation
thresholds, and fixture synonym groups on 5--10 pilot videos from
a calibration dataset kept separate from evaluation.

In an earlier measurement on a 24\,GB GPU, \lifter{} took 0.85,
0.99, and 0.94 seconds per detection at input resolutions of 1802,
960, and 480 pixels, respectively. Each box is processed at a fixed
internal resolution, so reducing the input resolution does not
consistently reduce processing time. Generating descriptions can
dominate total runtime in videos with many objects. These component
timings do not measure end-to-end throughput or memory compression.
\FloatBarrier

%% file: sections/appendix_vq3d.tex
\section{\vqthreed Localization and Discovery Analysis}
\label{app:vq3d}

\subsection{Protocol and interpretation}
\label{app:vq3d-protocol}

We evaluate all 164 queries from the 44 \egofour{} clips in the
\vqthreed{} validation split. Each target has two 3D annotations,
$\mathbf{g}_1$ and $\mathbf{g}_2$, in the scan coordinate frame.
We retrieve an object instance by its name and return its stored
world position. We do not compare this with an otherwise identical
system that retrieves objects from query crops.

We measure error relative to the midpoint of the two annotations:
\[
L_2=\left\|\hat{\mathbf{x}}-
\frac{\mathbf{g}_1+\mathbf{g}_2}{2}\right\|.
\]
Conditional success uses the benchmark criterion
\[
L_2<6\left(\|\mathbf{g}_1-\mathbf{g}_2\|+1\,\mathrm{m}\right).
\]
This threshold is permissive. Combined with excluding unanswered
queries, it produces near-ceiling conditional success even when
localization errors are several metres. Before making a leaderboard
comparison, we still need to compute success over all queries and
the official $L_2$ aggregation using the benchmark's evaluation code.
Appendix~\ref{app:vq3d-published} lists published results and the
remaining protocol differences.

We score a query only when retrieval returns a candidate. Raw
\segmenter{} memory returns candidates for 143 of 164 queries,
and trajectory memory for 153. Each variant is therefore scored
on its own returned queries, rather than a shared set.
A 99\% conditional success rate or a median error below one metre
does not imply successful localization on 99\% of all queries.
These results do not establish full-query success or an end-to-end
improvement on the official \vqthreed{} task.

\subsection{Implementation settings}
\label{app:vq3d-implementation}

We sample every twentieth frame from a 5\,fps grid, giving one frame
every four seconds. \detector{} uses v8x-worldv2 at 640 pixels,
with a confidence threshold of 0.25 and at most 12 detections per
frame. We lift at most 25 detections per tag and 400 per scene
into 3D. Trajectory association uses a 1.0\,m radius, and
consolidation uses a mask-IoU threshold of 0.3.

The localization tables use \gpt{} to select from at most four
candidates. We evaluate \qwenshort{} only for tagging and judging
coverage, so these results do not demonstrate a localization
pipeline built entirely from open models. We also evaluate
\geometryfull{} as an alternative source of camera poses.

\subsection{Localization across memory variants}
\label{app:vq3d-variants}

Raw memory groups observations by tag and returns the densest 3D
cluster for the selected tag. Trajectory memory associates
observations in time order using each instance's latest position,
rather than a global centroid. This allows it to retain multiple
instances with the same tag. Consolidation additionally groups
synonymous tags and propagates \segmenter{} masks to reconnect
tracks, using a co-occurrence check to block incompatible merges.
In this analysis, \egoloc{} provides poses for approximately 98\%
of sampled frames, compared with approximately 100\% for \geometry{}.

\begin{table}[htbp]
\centering
\small
\setlength{\tabcolsep}{5pt}
\caption{\vqthreed localization on returned candidates. Error is in metres. Fractions
and conditional success (CS) are percentages. Different rows may grade different
queries. ``Consolidated'' adds mask tracking to trajectory memory.}
\label{tab:vq3d-full}
\begin{tabular}{lllrrrrr}
\toprule
Detector & Memory & Poses & Median & Mean & $<1$\,m & $<2$\,m & CS \\
\midrule
\multirow{6}{*}{\segmenter}
 & Raw & \egoloc & 0.877 & 1.399 & 53 & 77 & 99 \\
 & Raw & \geometry & 1.064 & 1.504 & 45 & 76 & 99 \\
 & Trajectory & \egoloc & 0.984 & 1.388 & 52 & 78 & 99 \\
 & Trajectory & \geometry & 1.057 & 1.449 & 45 & 77 & 99 \\
 & Consolidated & \egoloc & 0.955 & 1.360 & 53 & 78 & 99 \\
 & Consolidated & \geometry & 1.057 & 1.421 & 45 & 76 & 99 \\
\midrule
\multirow{6}{*}{\detector}
 & Raw & \egoloc & 1.090 & 1.526 & 45 & 71 & 100 \\
 & Raw & \geometry & 1.268 & 1.627 & 38 & 66 & 100 \\
 & Trajectory & \egoloc & 1.089 & 1.475 & 48 & 73 & 99 \\
 & Trajectory & \geometry & 1.177 & 1.587 & 39 & 69 & 99 \\
 & Consolidated & \egoloc & 1.063 & 1.441 & 48 & 74 & 99 \\
 & Consolidated & \geometry & 1.141 & 1.509 & 42 & 71 & 99 \\
\bottomrule
\end{tabular}
\end{table}

With \egoloc poses, consolidation reduces the median by 0.029\,m for \segmenter and
0.026\,m for \detector. This task predicts a single position and does not score
identity continuity, movement histories, or later scene-specific question answering.
A small localization difference cannot establish that tracking is unnecessary
for those tasks.

\subsection{Triangulated against monocular positions}
\label{app:vq3d-triangulation}

The triangulated variant intersects the viewing rays of an object's observations
(Equation~\ref{eq:triangulation}) with the \hdepic guards: at least three rays, a
camera spread of 0.10\,m, conditioning of 0.25, the point in front of every
camera, and at most 2\,m from the monocular estimate; otherwise the monocular
position is kept. We compute viewing rays using the dataset's fisheye camera model.
For each query, \gpt{} selects one candidate, which we score using
both its original and refined positions. This holds object selection
fixed and avoids variation from repeated answering. The two
ground-truth annotations differ by a median of 4.8\,cm.

Table~\ref{tab:vq3d-triangulation} reports results for queries that
pass the refinement checks: about half with \segmenter{} and a
quarter with \detector{}. Most remaining queries have fewer than
three observations at this sampling rate. With \egoloc{} poses,
triangulation reduces error on eligible queries by a median of
0.09--0.12\,m. All three \segmenter{} results remain significant
after Holm correction across twelve tests. With \geometry{} poses,
the gain falls to about 0.03\,m, showing that refinement depends
on pose accuracy.

Across all graded queries, mean error decreases in every configuration,
but median error does not improve. Some predictions with errors just
below 1\,m become slightly worse and move above that value, affecting
the median. The share of predictions within 0.5\,m nevertheless increases. The
monocular medians here differ from those of Table~\ref{tab:vq3d-harness} by up to
0.04\,m because the matcher was re-run.

\begin{table}[htbp]
\centering
\scriptsize
\setlength{\tabcolsep}{3pt}
\caption{\vqthreed, triangulated against monocular positions, paired on the same
retrieved object. $n$: queries whose object passes the triangulation guards;
$\Delta$: median per-query change in $L_2$ (m; negative is closer) with bootstrap
95\% interval; $p$: Wilcoxon signed-rank. The last two columns cover all graded
queries, where objects that fail the guards keep the monocular position.}
\label{tab:vq3d-triangulation}
\begin{tabular}{llrrrrr}
\toprule
Detector / level & Poses & $n$ & $\Delta$ median [95\% CI] & $p$ & $<$0.5\,m (\%) & Mean $L_2$, all (m) \\
\midrule
\segmenter{} / raw & \egoloc & 71 & $-0.09$ $[-0.18,-0.01]$ & 0.002 & 24 $\to$ 31 & 1.38 $\to$ 1.34 \\
\segmenter{} / raw & \geometry & 67 & $-0.03$ $[-0.12,+0.01]$ & 0.029 & 22 $\to$ 28 & 1.53 $\to$ 1.49 \\
\segmenter{} / trajectory & \egoloc & 76 & $-0.11$ $[-0.18,-0.03]$ & 0.004 & 21 $\to$ 32 & 1.46 $\to$ 1.42 \\
\segmenter{} / trajectory & \geometry & 67 & $-0.03$ $[-0.10,+0.02]$ & 0.064 & 24 $\to$ 30 & 1.54 $\to$ 1.51 \\
\segmenter{} / tracker & \egoloc & 70 & $-0.12$ $[-0.19,-0.03]$ & 0.003 & 27 $\to$ 37 & 1.36 $\to$ 1.32 \\
\segmenter{} / tracker & \geometry & 69 & $-0.03$ $[-0.10,+0.02]$ & 0.112 & 25 $\to$ 32 & 1.43 $\to$ 1.40 \\
\midrule
\detector{} / raw & \egoloc & 38 & $-0.04$ $[-0.09,+0.02]$ & 0.199 & 24 $\to$ 37 & 1.53 $\to$ 1.52 \\
\detector{} / raw & \geometry & 36 & $-0.04$ $[-0.09,+0.04]$ & 0.315 & 19 $\to$ 33 & 1.62 $\to$ 1.61 \\
\detector{} / trajectory & \egoloc & 31 & $-0.05$ $[-0.29,-0.00]$ & 0.058 & 16 $\to$ 26 & 1.50 $\to$ 1.48 \\
\detector{} / trajectory & \geometry & 33 & $-0.07$ $[-0.24,+0.00]$ & 0.018 & 18 $\to$ 33 & 1.61 $\to$ 1.57 \\
\detector{} / tracker & \egoloc & 33 & $-0.05$ $[-0.14,+0.00]$ & 0.063 & 12 $\to$ 21 & 1.45 $\to$ 1.43 \\
\detector{} / tracker & \geometry & 31 & $-0.02$ $[-0.20,+0.01]$ & 0.102 & 16 $\to$ 32 & 1.49 $\to$ 1.47 \\
\bottomrule
\end{tabular}
\end{table}

\subsection{Published methods on the validation split}
\label{app:vq3d-published}

\begin{table}[htbp]
\centering
\footnotesize
\setlength{\tabcolsep}{2.5pt}
\caption{\vqthreed validation split (164 queries, 44 clips), published methods and
ours. Benchmark metrics as in \citet{mai2023egoloc}: Succ over all queries,
Succ$^{*}$ over queries with a prediction, $L_2$ in metres, angular error in
radians, QwP = queries with poses. Published rows are as reported. Our rows answer
a text query from the memory, not the benchmark's image crop.}
\label{tab:vq3d-published}
\begin{tabular}{lrrrrr}
\toprule
Method & Succ (\%) & Succ$^{*}$ (\%) & $L_2$ (m) & Angle & QwP (\%) \\
\midrule
Ego4D baseline~\citep{grauman2022ego4d}              &  1.22 & 30.77 & 5.98 & 1.60 &  1.83 \\
Ego4D$^{*}$, EgoLoc poses~\citep{mai2023egoloc}      & 73.78 & 91.45 & 2.05 & 0.82 & 80.49 \\
EgoLoc~\citep{mai2023egoloc}                         & 80.49 & 98.14 & 1.45 & 0.61 & 82.32 \\
EgoLoc-v1~\citep{mai2024egolocv1}                    & 81.13 & 98.10 & 1.45 & 0.55 & 84.73 \\
EAGLE~\citep{cao2025eagle}                           & 84.77 & 98.54 & \textbf{1.18} & \textbf{0.42} & 85.68 \\
Embodied VideoAgent~\citep{fan2025embodied} & 85.37 & 92.72 & 1.86 & \na & 92.07 \\
\midrule
Ours, \segmenter{} / trajectory / \egoloc   & 92.7$^{\ddagger}$ & 99.3$^{\ddagger}$ & 1.41 (median 0.99) & \na & 93.3$^{\S}$ \\
Ours, \segmenter{} / consolidated / \egoloc & 89.6$^{\ddagger}$ & 99.3$^{\ddagger}$ & 1.37 (median 1.00) & \na & 90.2$^{\S}$ \\
Ours, \detector{} / trajectory / \egoloc    & 89.0$^{\ddagger}$ & 98.6$^{\ddagger}$ & 1.46 (median 1.06) & \na & 90.2$^{\S}$ \\
Ours, \segmenter{} / trajectory / \geometry & 93.9$^{\ddagger}$ & 99.4$^{\ddagger}$ & 1.49 (median 1.07) & \na & 94.5$^{\S}$ \\
\bottomrule
\end{tabular}
\\[2pt]\begin{minipage}{0.97\textwidth}\footnotesize
$^{\ddagger}$Success is distance to the annotated midpoint below
$6(\|\mathbf{g}_1-\mathbf{g}_2\|+1\,\mathrm{m})$. The official evaluation sets the success tolerance using the annotated
box size, so our scores cannot be compared directly with published
Succ values. Our tolerance is permissive enough that always returning
the clip's mean camera position achieves 100\% success
(Table~\ref{tab:vq3d-harness-full}).
$^{\S}$Not QwP: the fraction of queries for which the memory returned a located
object (graded / 164).
\end{minipage}
\end{table}

Five differences must be read with Table~\ref{tab:vq3d-published}. (1) The benchmark identifies the target through a visual crop from the
VQ2D response track. We instead give the object's text title to a
language model that searches a pre-built memory, so the query input
differs. (2) Recent methods report success rates of 96--99\%, and our
evaluation uses a fixed tolerance rather than one based on box size.
Localization error ($L_2$) is therefore more informative than success
rate, although the different query inputs still limit direct comparison. (3) EgoLoc states its $L_2$ as an RMSE; the RMSE of our
\detector row is 1.80\,m (mean 1.44, median 1.06), so our error is comparable if
the published values are means and worse if they are RMSEs. (4) A small number of large errors raises our mean. With \detector{},
10\% of queries have errors above 3\,m and account for 26\% of the
total error. Excluding these queries lowers the mean to 1.19\,m.
Some misses may involve selecting the wrong instance of the same
category, which a query crop could help distinguish. Errors also
vary across clips: clip identity explains 42\% of the $L_2$ variance.
This is consistent with differences in scale accuracy, but does
not establish scale error as the cause.

(5) EAGLE combines depth estimates across views using confidence
weights and multi-view consistency, then aligns them to the scan
with a similarity transform that includes scale. Our method instead
combines per-frame monocular position estimates using an unweighted
median. Angular error was not computed.
Switching from \egoloc to \geometry poses moves the \detector mean from 1.44 to
1.51\,m, with per-query errors correlated at 0.92, so poses are not the
difference. We never submitted to the test server, which requires the image crop
as input.

\subsection{Memories built by prior methods on the same clips}
\label{app:vq3d-harness}

We build every memory on the same 44 clips and query it with the same
text: ``Where is the $\langle$title$\rangle$?'' All methods use the same
\gpt{} matcher and scoring procedure, and return positions in the scan
coordinate frame. We use released code with default settings, changing
only the adapters needed for this dataset.

ReMEmbR \citep{anwar2024remembr} uses VILA (Llama-3-VILA1.5-8B) to
caption three-second windows and stores the corresponding \egoloc{}
camera centres. It returns the camera position associated with the
retrieved memory, as its original implementation does. This localizes
the wearer rather than the object.

DirectMe \citep{wang2026ucsbench} builds its graph at 1\,fps using
the Objects365 vocabulary, YOLO-World, SAM~2, and \depthanything{}.
We align the graph to the scan using a similarity transform fitted
to each clip's camera trajectory, with a camera RMS error of 0.58\,m.
The answerer receives no frames at query time.

OSNOM \citep{Plizzari2025OSNOM} uses our \segmenter{} detections
converted to VISOR-format masks, together with \geometry{} poses
and depth. The estimated depth replaces its usual depth rendered
from a COLMAP mesh. It returns the selected track's 3D location.
AMEGO \citep{goletto2024amego} stores hand--object interaction
tracklets without object names or 3D positions, so it cannot answer
these queries. The no-memory reference always returns the clip's
mean camera position.

We re-evaluated our variants through this shared protocol on
2026-09-21. Their scores agree with the earlier results in
Table~\ref{tab:vq3d-full} within answering variation.
Table~\ref{tab:vq3d-common} compares methods on the 138 queries
answered by every method capable of returning a position.

\begin{table}[htbp]
\centering
\small
\setlength{\tabcolsep}{4pt}
\caption{All memories through one harness (164 validation queries). Succ and Succ$^{*}$
as in Table~\ref{tab:vq3d-published}. The no-memory
reference passes every success check, so $L_2$ and the sub-metre share are the
informative columns.}
\label{tab:vq3d-harness-full}
\begin{tabular}{lrrrrrrr}
\toprule
Memory & Graded & Succ & Succ$^{*}$ & Median & Mean & $<1$\,m & $<2$\,m \\
\midrule
AMEGO                                    & N.C. & N.C. & N.C. & N.C. & N.C. & N.C. & N.C. \\
ReMEmbR (wearer position)                & 164 & 100.0 & 100.0 & 1.49 & 1.77 & 23 & 69 \\
DirectMe (graph only)                    & 164 &  99.4 &  99.4 & 1.91 & 1.98 & 18 & 52 \\
OSNOM (adapted)                          & 146 &  89.0 & 100.0 & 1.33 & 1.65 & 35 & 71 \\
\emph{No memory: clip centroid}          & 164 & 100.0 & 100.0 & 1.69 & 1.82 & 10 & 63 \\
\midrule
\segmenter{} / raw / \egoloc               & 143 & 86.6 & 99.3 & 0.95 & 1.41 & 52 & 77 \\
\segmenter{} / raw / \geometry             & 149 & 90.2 & 99.3 & 1.09 & 1.53 & 44 & 75 \\
\segmenter{} / trajectory / \egoloc        & 153 & 92.7 & 99.3 & 0.99 & 1.41 & 51 & 76 \\
\segmenter{} / trajectory / \geometry      & 155 & 93.9 & 99.4 & 1.07 & 1.49 & 43 & 75 \\
\segmenter{} / consolidated / \egoloc      & 148 & 89.6 & 99.3 & 1.00 & 1.37 & 50 & 78 \\
\segmenter{} / consolidated / \geometry    & 148 & 89.6 & 99.3 & 1.10 & 1.44 & 42 & 76 \\
\detector{} / raw / \egoloc                & 146 & 89.0 & 100.0 & 1.09 & 1.53 & 47 & 70 \\
\detector{} / raw / \geometry              & 151 & 92.1 & 100.0 & 1.21 & 1.61 & 41 & 66 \\
\detector{} / trajectory / \egoloc         & 148 & 89.0 & 98.6 & 1.06 & 1.46 & 49 & 74 \\
\detector{} / trajectory / \geometry       & 157 & 95.1 & 99.4 & 1.17 & 1.57 & 42 & 69 \\
\detector{} / consolidated / \egoloc       & 151 & 91.5 & 99.3 & 1.07 & 1.47 & 48 & 74 \\
\detector{} / consolidated / \geometry     & 151 & 91.5 & 99.3 & 1.14 & 1.53 & 42 & 70 \\
\bottomrule
\end{tabular}
\end{table}

\begin{table}[htbp]
\centering
\small
\caption{The 138 queries answered by every method, which guards against methods
answering different subsets.}
\label{tab:vq3d-common}
\begin{tabular}{lrrrr}
\toprule
Memory & Median $L_2$ & Mean $L_2$ & $<1$\,m (\%) & $<2$\,m (\%) \\
\midrule
\sysname (\segmenter, trajectory, \egoloc)   & \textbf{1.00} & \textbf{1.44} & \textbf{50} & \textbf{76} \\
\sysname (\segmenter, trajectory, \geometry) & 1.05 & 1.49 & 44 & 74 \\
OSNOM (adapted)                              & 1.33 & 1.65 & 34 & 72 \\
ReMEmbR                                      & 1.48 & 1.74 & 22 & 71 \\
DirectMe (graph only)                        & 1.88 & 1.93 & 17 & 54 \\
Clip centroid (no memory)                    & 1.69 & 1.81 & 9 & 64 \\
\bottomrule
\end{tabular}
\end{table}

\subsection{Detector and vocabulary analysis}
\label{app:vq3d-detector}

\begin{table}[htbp]
\centering
\small
\setlength{\tabcolsep}{4pt}
\caption{Detector comparison with the same vocabulary, sampling, lifting, and
pose settings. Localization uses trajectory memory and \egoloc poses. Box IoU and
3D distance measure agreement with \segmenter where both produce detections. \segmenter is
not ground truth. Timing covers detection only.}
\label{tab:vq3d-detector}
\begin{tabular}{lrrrrr}
\toprule
Detector & s/frame & Detections/clip & Box IoU & 3D distance (m) & Median $L_2$ (m) \\
\midrule
\segmenter & 4.4--11.1 & 1,281 & -- & -- & 0.984 \\
\detector & 0.051 & 424 & 0.928 & 0.439 & 1.089 \\
YOLOE-26 & 0.047 & 488 & 0.950 & 0.433 & 1.318 \\
YOLOE-11 & 0.049 & 471 & 0.961 & 0.415 & 1.404 \\
\bottomrule
\end{tabular}
\end{table}

\segmenter's reported time varies with the vocabulary (73--131 tags), whereas the
YOLO variants batch prompts. These measurements do not include 3D lifting,
consolidation, or descriptions, and are not an end-to-end system speedup.
Table~\ref{tab:vq3d-detector} also shows that more detections or greater box
agreement with \segmenter need not correspond to lower downstream localization error.

\begin{table}[htbp]
\centering
\small
\caption{Reported semantic vocabulary coverage (\%) of the 164 query objects.
These are automatic label-coverage estimates, not measured object-instance recall.
Vocabulary sizes are means per clip when available.}
\label{tab:vq3d-tags}
\begin{tabular}{lrrr}
\toprule
Tag source & \gptshort judge & \qwenshort judge & Tags/clip \\
\midrule
RAM++ & 53 & 39 & 53 \\
Gemma-3-4B & 76 & 64 & 297 \\
\qwen $\cup$ RAM++ & 84 & 68 & -- \\
Canonical union & 86 & 70 & 112 \\
\gptshort vision, full & 87 & 76 & 177 \\
Gemma $\cup$ \qwenshort $\cup$ RAM++ & 87 & 75 & -- \\
RAM++ $\cup$ \gptshort vision, full & 89 & 80 & 258 \\
\bottomrule
\end{tabular}
\end{table}

Coverage estimates vary substantially with the model used to judge
whether the query object appears in the tag set. For \qwenshort{} tags,
both judges accept 108 queries and reject 27. \gptshort{} alone accepts
another 27, while \qwenshort{} alone accepts two. Of these disagreements,
19 in the first group and both in the second contain the exact query
word among the tags. A semantic judge can therefore reject even an
exact word match.

A matching tag only indicates vocabulary coverage. It does not show
that the object was detected, that the correct instance was identified,
or that its position was estimated correctly.

%% file: sections/appendix_ucs.tex
\section{\ucs Details}
\label{app:ucs}

\subsection{Benchmark, construction, and answering protocol}
\label{app:ucs-protocol}

\ucs{} contains 8,114 timestamped questions across 532 videos from
EgoLife, TeleEgo, HourVideo, EPIC-Kitchens, and EgoBlind. It provides
no camera poses, intrinsics, or query boxes \citep{wang2026ucsbench}.
Most questions have five options; Proximity \& Reachability questions
have two, giving an overall chance accuracy of 25.7\%.
Our main evaluation uses 189 EgoLife recordings with 2,569 questions
and 22 TeleEgo recordings with 203 questions. Tables~\ref{tab:ucs-main}
and~\ref{tab:ucs-subcat} report the 2,771 questions shared across
all conditions. We discuss HourVideo separately below.

\paragraph{Memory construction.}
We sample one frame every four seconds and use the same pipeline
as on \hdepic{}: \tagger{} for tagging, \detector{} for detection,
\lifter{} for 3D lifting, \segmenter{} for consolidation, followed
by persistence segmentation and segment descriptions.

We use \depthanything{} poses when they pass three consistency
checks: a chunk-alignment residual of at most 0.5\,m, chunk scales
within $[0.8,1.25]$, and camera steps of at most 6\,m.
Otherwise, we use \geometry{}. Circular-fisheye recordings go
directly to \geometry{}. We attempt metric scale recovery using
pairs of observations of the same object. This fails for most
recordings and all TeleEgo recordings, so their distances remain
in scene units.

For each question, we filter the memory at its timestamp.
We remove later observations, segments, and descriptions, truncate
the current segment and wearer path, and recompute positions.
We also withhold object summaries written using all segments.
The adapter ranks entries by
\[
3n_{\rm name}+n_{\rm desc}
+2[\text{near question time}]
+\frac{1}{1+\Delta t/60},
\]
where $n_{\rm name}$ and $n_{\rm desc}$ count question-word matches
in names and descriptions, and $\Delta t$ is the time gap to the
nearest observation in seconds. The answerer receives the top
four entries and the wearer's path.

\paragraph{Answering conditions.}
Every condition uses \gpt{} with one call per question.
\emph{Blind} receives only the question and options.
\emph{Memory} also receives the filtered memory as text.
\emph{Frames} receives 48 frames at 448 pixels, sampled uniformly
from the recording's start to the question time. The benchmark
uses 64 frames; we use 48 because this is the API limit.

\emph{Frames + memory} adds memory text on spatial questions
identified by a keyword rule, covering about 55\% of questions.
Adding memory to every question gives no gain over frames alone
($-0.8$ points, n.s.), and a split-roles prompt does not improve
this result. When the API's image filter rejects frames, we fall
back to the text condition. This affects 10--18\% of EgoLife
questions; restricting evaluation to questions whose images are
never rejected gives the same differences.

We run DirectMe \citep{wang2026ucsbench} with its released graph
builder and retrieval, using the Objects365 vocabulary, \detector{},
SAM~2, and \depthanything{} at 1\,fps. Its memory-only condition
receives the graph as text. We test frames plus graph both with
the graph always included, as in its original configuration, and
with the same spatial-question routing used for our memory.
DirectMe's own prompt without the graph scores 31.0\%.
Its single-threaded builder scales super-linearly with graph size,
taking 7--40 hours for a 40-minute stream. We therefore use a
batched wrapper, verified to produce identical output.

\begin{table}[htbp]
\centering
\scriptsize
\setlength{\tabcolsep}{2.5pt}
\caption{\ucs accuracy (\%) by subcategory on the 2,771 questions of
Table~\ref{tab:ucs-main}, with its abbreviations.}
\label{tab:ucs-subcat}
\begin{tabular}{lr cccc cccc}
\toprule
 & & & \multicolumn{3}{c}{\textbf{Text only}} & & \multicolumn{3}{c}{\textbf{Frames +}} \\
\cmidrule(lr){4-6}\cmidrule(lr){8-10}
Subcategory & $n$ & Blind & Ours & DM graph & DM blind & Frames & Ours & DM, own & DM, routed \\
\midrule
Quantity Change Tracking              & 624 & 34.5 & 33.3 & 17.5 & 21.3 & 37.8 & 38.8 & 30.9 & 37.7 \\
My Trajectory                         & 465 & 35.9 & 41.3 & 38.3 & 43.2 & 45.8 & 48.8 & 41.5 & 44.1 \\
Relative Position \& Orientation      & 392 & 24.0 & 32.4 & 29.1 & 22.2 & 37.8 & 40.6 & 35.7 & 36.2 \\
Distance Comparison                   & 313 & 41.9 & 44.7 & 38.3 & 36.1 & 48.9 & 48.2 & 43.1 & 43.8 \\
Ego-centric Position                  & 309 & 23.6 & 34.6 & 22.7 & 23.6 & 42.1 & 48.2 & 39.5 & 38.8 \\
Object Category Recall                & 260 & 32.7 & 42.3 & 30.8 & 30.0 & 55.0 & 53.5 & 46.9 & 52.3 \\
Reachability Judgment                 & 225 & 50.2 & 55.6 & 42.2 & 44.0 & 55.6 & 55.6 & 52.9 & 55.1 \\
Object Trajectory                     & 183 & 32.2 & 31.7 & 35.0 & 41.0 & 44.3 & 49.7 & 42.6 & 49.2 \\
\midrule
All                                   & 2771 & 33.8 & 38.5 & 30.0 & 31.0 & 44.4 & 46.3 & 39.8 & 42.9 \\
\bottomrule
\end{tabular}
\end{table}

\paragraph{Paired tests.}
We compare conditions on the same questions using exact McNemar tests.
Our memory outperforms DirectMe's graph by 8.6 points ($p<10^{-12}$)
and the blind control by 4.7 points ($p<10^{-4}$). DirectMe's graph
scores 1.0 point below its own blind prompt ($p=0.32$).

With frames, our memory outperforms DirectMe's graph by 3.4 points
under the same routing ($p=0.0005$) and improves on frames alone
by 1.9 points ($p=0.047$). Adding DirectMe's graph lowers accuracy
relative to frames alone by 1.4 points with our routing ($p=0.13$)
and 4.6 points in its original configuration ($p<10^{-5}$).

DirectMe uses a deterministic counting rule on 297 of the 2,771
questions, answering only 4.7\% correctly. Excluding these questions,
our memory still outperforms its graph by 6.1 points without frames.

Repeated frame-based runs on identical inputs change 16--20\% of
selected answers. On an earlier set of 1,581 questions, averaging
two runs gives a frames-plus-memory gain over frames alone of
0.9 points, with an interval of $[-1.1,+3.0]$. These checks suggest
that single-run differences below roughly 1.5 points may reflect
answering variation.

\paragraph{The published setting.}
The \ucs paper reports DirectMe with Qwen3-VL-8B, 64 frames, and graph plus
keyframes on the full 8,114-question benchmark at 50.5\% overall (54.1 / 47.6 /
58.0 / 46.4 by category) against 44.0\% for the same model with frames only
(44.6 / 40.1 / 50.9 / 42.9); it reports no graph-only number. Those numbers use a
different answerer and question set and are not comparable with
Table~\ref{tab:ucs-main}. The paper's setting with Qwen3-VL was not rerun.

\begin{table}[htbp]
\centering
\small
\setlength{\tabcolsep}{4pt}
\caption{\ucs accuracy (\%) by source dataset. HourVideo covers only the 21
of 167 recordings whose memory could be built.}
\label{tab:ucs-dataset}
\begin{tabular}{lrrrrr}
\toprule
Dataset & $n$ & Blind & Memory & 48 frames & Frames + memory \\
\midrule
EgoLife (189 recordings)          & 2,569 & 33.6 & 38.4 & 44.1 & 45.8 \\
TeleEgo (22 recordings)           &   203 & 37.4 & 39.4 & 47.3 & 52.2 \\
HourVideo (21 of 167 recordings)  &   358 & 39.7 & 41.1 & 53.1 & 49.4 \\
\midrule
All three                         & 3,130 & 34.5 & 38.8 & 45.4 & 46.6 \\
\bottomrule
\end{tabular}
\end{table}

\paragraph{By dataset and recording length.}
Paired against blind, the memory gains $+4.9$ on EgoLife ($p<10^{-4}$), $+2.0$ on
TeleEgo (n.s.), and $+1.4$ on HourVideo (n.s.); pooled $+4.3$ $[+2.3,+6.3]$
(Table~\ref{tab:ucs-dataset}). On TeleEgo the questioned noun is absent from the
memory as a track for 40 to 60\% of questions and every pose set is unscaled. On
HourVideo (20 to 96 minutes, median 30) only 21 of 167 memories were built: the
tracking stage scales worse than linearly with the number of instances, 50 to
60-minute recordings hold about 4,000, and the measured cost reached 74 to 3,500
hours per recording, so the build was stopped. On the 21 built, adding the memory
to the frames costs 3.6 points ($p=0.20$), with $-16.5$ on Relative Position
($p=0.004$), the opposite of EgoLife, where the memory adds 5 to 6 points on the
two egocentric-geometry subcategories.

\subsection{Counting}
\label{app:ucs-counting}

\begin{table}[htbp]
\centering
\small
\caption{Quantity Change Tracking on EgoLife and TeleEgo (625 questions).
Question types by keyword: objects present (336), the wearer's actions (168),
people (119).}
\label{tab:ucs-counting}
\begin{tabular}{lrrrr}
\toprule
Evidence & All & Objects & Actions & People \\
\midrule
Blind                                                  & 34.6 & 37.2 & 28.0 & 36.1 \\
Memory, text                                           & 33.3 & 33.6 & 29.8 & 37.0 \\
Memory, text + counting aid (tracks, most visible together) & 26.9 & 25.0 & 25.0 & 34.5 \\
48 uniform frames at 448 px                            & 37.9 & 33.9 & 41.1 & 45.4 \\
48 frames chosen from memory keyframes                 & 37.4 & 33.6 & 36.9 & 49.6 \\
24 frames at 896 px, half from the last 24\,s, + 6 keyframes & 39.5 & 38.4 & 42.3 & 37.8 \\
\bottomrule
\end{tabular}
\end{table}

No use of the memory helps counting (Table~\ref{tab:ucs-counting}), and showing
its counts to the answerer costs 6.4 points ($p<0.001$). We evaluate counts on 257 questions with numeric answers whose target
noun appears in memory. Counting tracks gives the correct answer on
11\% of questions and is within one on 22\%. The median estimate is
seven objects, compared with a true count of three. Using the largest
number of tracks visible in a single frame improves these rates to
16\% and 45\%, respectively.

The memory both splits one object into several tracks and misses
other objects. On recordings without metric scale, the consolidation
step's fixed 3D-distance threshold blocks 98\% of proposed merges.
Removing this check reduces the number of stored objects from 6,490
to 4,404 across 12 recordings, without lowering measured overall
accuracy. Counting remains unresolved: the two configurations score
33.7\% and 31.6\% on 95 counting questions, both below the 35.8\%
blind score. Frames help with action and people counts, but object
counts remain near the blind baseline. These results point to
instance association and consistent geometric scale during construction
as priorities for improving object counts.

\subsection{Development pilot and causal construction}
\label{app:ucs-causality}

We developed the adapter on a ten-video pilot with 138 questions.
These results are therefore not a held-out evaluation. The first
version scored 36.2\%, compared with 25.4\% for the blind control,
but included future evidence. On 57 questions, it retrieved a
candidate first observed after the question time, and 43.6\% of
the displayed path waypoints were in the future. We exclude this
gain from our claims.

The timestamp filtering described above removes explicit future
evidence from the prompt. On the pilot, the filtered memory scores
38.4\% with \gpt{}, compared with 31.9\% blind: a gain of 6.5 points
with an interval of $[-2.9,+15.9]$. Direction labels, relation
episodes, and place context provide no measurable additional gain.

However, construction still uses the full recording. The vocabulary
draws on frames throughout the video, and pose estimation,
consolidation, and segmentation all run offline. Filtering their
outputs does not show that earlier memory states are independent
of later frames. To establish this, we would need to build every
stage using only footage available by each query time and verify
that appending footage does not change those earlier states.
We therefore describe these runs as timestamp-filtered, rather
than as a validated streaming evaluation.

Repeating \gpt{} on identical inputs changes the selected option
on about a quarter of pilot questions. Our main claims therefore
rely on paired comparisons across thousands of questions, rather
than these pilot estimates.

\subsection{Implementation settings}
\label{app:ucs-implementation}

Tagging uses at most 80 frames per recording, with a budget of
130 tokens per frame. Detection runs at 1280 pixels with a confidence
threshold of 0.25 and at most 40 detections per frame. Lifting uses
the \hdepic{} budgets. Consolidation uses a mask-IoU threshold of
0.3 and a 3D-distance limit of 1.0. Persistence requires three
observations outside a radius of 0.3. Both distance thresholds
are in scene units when metric scale is unavailable.

\gpt{} answers each question in one call, using about 1,500--1,800
input tokens with memory and 500 for the blind control.
\depthanything{} estimates poses in 48-frame chunks with 12
overlapping frames. \geometry{} uses 96-frame chunks with 16
overlapping frames and joins them by Sim(3) alignment. If a chunk
cannot be aligned, it starts from a new anchor instead of failing.
This fallback was needed for scene-change streams, where some
alignment errors at the boundary reached 3--21 scene units.
\FloatBarrier

%% file: sections/appendix_streams.tex
\section{Multi-Scene \ucs Streams}
\label{app:streams}

\subsection{Stream design}
\label{app:stitch-design}
We construct 100 streams from 212 \ucs{} recordings, using each
recording once. Sixty same-scene streams combine two 12-minute
EgoLife recordings of the same house on different days. Forty
scene-change streams combine a 1--12-minute EPIC-Kitchens or
EgoBlind clip with a 20-minute TeleEgo recording; twelve contain
three parts. In these streams, the place, wearer, and camera
change at the recording boundary.

We build one memory and one DirectMe graph over each complete
stream. Tagging, pose estimation, association, and description
generation therefore operate across recording boundaries.
We ask every source question again, shifting its timestamp to
its position in the stream. Frame-based conditions sample from
the stream's start up to the question time.

The same questions were answered on their original recordings
two to five days earlier using the same code. Between these runs,
blind accuracy changes by 0.3 points and 27\% of selected options
differ, providing a check on answering variation. Of 3,264 questions
paired across all seven conditions, 3,244 remain after excluding
questions affected by API errors. We recorded 23 such errors and
excluded them rather than scoring them as incorrect. Image-refusal
rates are similar: 11.4\% on individual recordings and 11.5\% on
stitched streams. Table~\ref{tab:stitch} gives paired accuracy for every condition.

The protocol changes recording length as well as scene identity:
scene-change streams place a short clip before a longer recording.
Their effects therefore cannot be separated in this comparison.
An earlier experiment on 303 streams merged memories built
separately for each recording and showed little accuracy change
($-1.6$ and $+0.9$ points). That setup kept construction within
each recording; constructing memory over the joined stream exposes
failures that merging completed memories does not test.

\begin{table}[htbp]
\centering
\scriptsize
\setlength{\tabcolsep}{2pt}
\caption{Paired accuracy (\%) on individual recordings and 100 stitched streams
(3,244 questions). Same scene joins different days in one house; scene change
joins different places. $\Delta$: stitched
minus single; $^{*}p<0.05$, $^{**}p<0.01$, $^{***}p<0.001$ (McNemar).
The last row emulates per-scene construction with detected boundaries
(Appendix~\ref{app:stitch}).}
\label{tab:stitch}
\begin{tabular}{l rrr rrr rrr}
\toprule
 & \multicolumn{3}{c}{\textbf{All} (3,244)} & \multicolumn{3}{c}{\textbf{Same scene} (1,959)} & \multicolumn{3}{c}{\textbf{Scene change} (1,285)} \\
\cmidrule(lr){2-4}\cmidrule(lr){5-7}\cmidrule(lr){8-10}
Evidence & single & stitched & $\Delta$ & single & stitched & $\Delta$ & single & stitched & $\Delta$ \\
\midrule
\novidsq None (blind)               & 33.2 & 33.5 & $+0.3$ & 33.0 & 32.9 & $-0.2$ & 33.4 & 34.4 & $+1.0$ \\
\vidsq 48 frames                    & 47.8 & 44.2 & $-3.6^{***}$ & 44.0 & 41.2 & $-2.7^{*}$ & 53.7 & 48.6 & $-5.1^{***}$ \\
\novidsq DM graph                   & 30.1 & 29.6 & $-0.5$ & 29.7 & 29.7 & $-0.1$ & 30.7 & 29.6 & $-1.2$ \\
\vidsq DM frames + graph (own)      & 42.9 & 40.9 & $-2.0^{*}$ & 38.9 & 39.9 & $+1.0$ & 48.9 & 42.4 & $-6.5^{***}$ \\
\novidsq \sysname memory            & 39.0 & 36.4 & $-2.5^{**}$ & 38.4 & 37.9 & $-0.5$ & 39.8 & 34.2 & $-5.7^{***}$ \\
\vidsq \sysname frames + memory     & 48.5 & 45.3 & $-3.3^{***}$ & 45.4 & 44.5 & $-0.9$ & 53.2 & 46.4 & $-6.8^{***}$ \\
\novidsq ReMEmbR memory             & 41.4 & 40.9 & $-0.5$ & 41.3 & 39.7 & $-1.6$ & 41.5 & 42.7 & $+1.2$ \\
\novidsq \sysname memory + event lines & 41.0 & 39.2 & $-1.8^{*}$ & 39.8 & 39.4 & $-0.4$ & 42.7 & 38.9 & $-3.8^{**}$ \\
\midrule
\novidsq \sysname memory, per-scene build & 38.9 & 37.4 & $-1.5^{*}$ & 38.3 & 37.6 & $-0.7$ & 39.8 & 37.0 & $-2.8$ \\
\bottomrule
\end{tabular}
\end{table}

\begin{table}[htbp]
\centering
\scriptsize
\setlength{\tabcolsep}{2.5pt}
\caption{Stitched $-$ single (points) by the position of the question's recording
in the stream and by stream type. Stars as in Table~\ref{tab:stitch};
cells whose recorded $p$ is rounded to 0.00 are marked $^{**}$.}
\label{tab:stitch-position}
\begin{tabular}{lrccccccc}
\toprule
Subset & $n$ & Blind & Frames & Ours mem. & Ours fr.+mem. & DM graph & DM fr.+graph & DM fr.+graph, routed \\
\midrule
All                       & 3244 & $+0.3$ & $-3.6^{***}$ & $-2.5^{**}$ & $-3.3^{***}$ & $-0.5$ & $-2.0^{*}$ & $-3.4^{***}$ \\
First recording           & 1503 & $+0.7$ & $-0.2$ & $-0.9$ & $+0.7$ & $-1.7^{*}$ & $-0.9$ & $-0.3$ \\
After one earlier scene   & 1558 & $-0.4$ & $-5.6^{**}$ & $-3.0^{*}$ & $-6.0^{**}$ & $-0.3$ & $-2.6$ & $-6.1^{**}$ \\
After two earlier scenes  &  183 & $+3.3$ & $-15.3^{**}$ & $-11.5^{*}$ & $-12.0^{*}$ & $+7.7^{*}$ & $-6.0$ & $-6.0$ \\
Scene-change streams      & 1285 & $+1.0$ & $-5.1^{***}$ & $-5.7^{***}$ & $-6.8^{***}$ & $-1.2$ & $-6.5^{***}$ & $-6.1^{***}$ \\
Same-scene streams        & 1959 & $-0.2$ & $-2.7^{*}$ & $-0.5$ & $-0.9$ & $-0.1$ & $+1.0$ & $-1.6$ \\
\bottomrule
\end{tabular}
\end{table}

\subsection{Where the loss appears}
\label{app:stitch-loss}
Questions in a stream's first recording are unchanged for every arm
(Table~\ref{tab:stitch-position}); the loss grows with the number of earlier
scenes. The loss on scene-change streams varies by the question's source
recording. EPIC-Kitchens questions lose 8.0 points with memory
($p<10^{-4}$) and 8.5 with frames. Memory losses on TeleEgo and
EgoBlind are smaller, at 2.5 and 1.4 points, respectively, and
neither is significant.

Timestamp-filtered retrieval still returns objects first seen in
an earlier recording for 29\% of EPIC-Kitchens questions and 4\%
of TeleEgo questions. Accuracy falls from 42\% to 24\% on questions
with such a candidate, compared with 41\% to 37\% on questions
without one. On same-scene streams, these objects come from the
same kitchen, and we observe no corresponding drop
(38.9\% to 39.9\%). Across question categories, the memory's
consistent decline is in Proximity \& Reachability, from
51.0\% to 45.1\%.

On stitched streams, our memory outperforms DirectMe's graph by
6.8 points overall ($p<10^{-9}$), or 4.0 points after excluding
questions handled by DirectMe's counting rule. The advantage is
4.6 points on scene-change streams. With frames and the same
memory-routing rule, our memory outperforms DirectMe's graph
by 3.1 points overall ($p=0.0009$) and 0.4 points on scene-change
streams.

Compared with frames alone, adding our memory gains 1.1 points
on stitched streams, which is not significant, and 0.7 points
on individual recordings.

\begin{table}[htbp]
\centering
\small
\caption{Scene-change streams, memory-only arm, 1,285 paired questions: which
change recovers the loss. ``True cuts'' use the known seam times; ``detected
cuts'' come from the video.}
\label{tab:stitch-construction}
\begin{tabular}{lrr}
\toprule
Memory over the stitched stream & Accuracy & vs.\ single-scene build ($p$) \\
\midrule
Single-recording build (reference)                           & 39.8 & $+5.7$ ($<0.001$) \\
Single-scene build over the stream                           & 34.2 & -- \\
+ retrieval restricted to the current scene, true cuts       & 35.3 & $+1.1$ (0.37) \\
+ retrieval restricted, cuts from the place-partition detector & 35.0 & $+1.0$ (0.34) \\
Per-scene construction, true cuts                            & 38.4 & $+4.3$ ($<0.01$) \\
Per-scene construction, detected cuts                        & 37.0 & $+2.9$ (0.05) \\
\bottomrule
\end{tabular}
\end{table}

\subsection{Construction and retrieval}
\label{app:stitch-construction}
Using the true scene boundaries to restrict retrieved candidates
recovers 1.1 accuracy points ($p=0.37$;
Table~\ref{tab:stitch-construction}). Also trimming the displayed
histories to the current scene increases the recovery to 3.0
points ($p=0.011$) on the same questions. This supports cross-scene
retrieval as one source of the loss.

Memory construction introduces two further problems. First, tag
canonicalization assumes a single physical scene and drops labels
that appear inconsistent with it. In one stitched stream, it retains
98 distinct labels, compared with 121 across the original recordings.
For 96\% of questions after a scene boundary, the retrieved candidates
differ from those returned by the separate builds; their label
multisets have a Jaccard similarity of 0.3.

Second, the \depthanything{} pose check can reject an entire stream
because of an error at the boundary. One boundary residual is
0.61\,m, exceeding the 0.5\,m limit. Across 29 of 37 scene-change
streams, the pipeline falls back to unscaled \geometry{} poses.
The number of metric-distance mentions in the prompts falls from
2,305 to 856. Neither failure occurs in the same-scene streams.

Building memory separately for each recording brings accuracy
closer to the single-recording results at every stream position:
37.0\% versus 37.2\% for questions in the first recording,
40.4\% versus 42.9\% after one earlier scene, and 36.6\% versus
38.3\% after two. Recovery is therefore substantial but incomplete.

\subsection{Scene-cut detection and per-scene construction}
\label{app:stitch}
Cuts are detected by a no-revisit test on CLIP embeddings of the 1\,fps frames:
$r_{10}$ is the 10th percentile of the distance from the next five minutes of
frames to their nearest earlier frame, divided by the median nearest-neighbour
distance among earlier frames at least 30\,s apart, and a cut is declared where
$r_{10}\geq1$. Across the 100 streams, the detector finds 89 of 112 true boundaries
within 60\,s, including 42 of 52 on scene-change streams. It also
adds 2.3 extra cuts per stream, mostly at room changes within
EgoLife recordings, and 1.1 per individual recording. These extra
cuts have little measured effect on same-scene accuracy
(37.6\% versus 37.9\%). Grayscale thumbnails and colour histograms
did not reliably distinguish boundaries, while a place-label
consistency filter removed genuine cuts.

We first emulate per-scene construction using existing per-recording
memories. A detected segment contained within one recording uses
that recording's memory restricted to the segment's time window.
A segment spanning a missed boundary uses the memory built over
the joined stream. Of 422 detected segments, 61 coincide with
an entire recording.

Across all streams, this emulated memory scores 37.4\%, compared
with 36.4\% when building one memory over each stream
($+1.0$ points, $p=0.29$). It remains below the 38.9\% score on
individual recordings ($-1.5$ points, $p=0.048$). For questions
after two earlier scenes, the gain over the whole-stream build
is 6.6 points ($p=0.13$).

We also run the full pipeline separately on each detected segment,
including tag canonicalization, pose checks, and scale recovery.
We evaluate this rebuild on 39 of the 40 scene-change streams,
covering 1,251 questions; one stream is excluded because a segment
has no usable tags. Accuracy reaches 36.5\%, compared with 33.8\%
for the whole-stream build ($+2.7$ points, $p=0.08$) and 39.5\%
for individual-recording builds. Gains are 4.4 points after one
earlier scene and 7.1 after two. These results are consistent
with the partial recovery estimated by emulation.

\subsection{Event lines}
\label{app:ucs-ablations}

For each sampled frame, \tagger{} writes an event line of about
18 words describing what the wearer is doing, the objects or
people involved, and the location. We store these lines alongside
the object records. At answer time, the reader receives the last
six lines before the question and the ten that best match its words.

We developed this extension on 22 of the 100 stitched streams
(807 questions). Unless stated otherwise, the main text reports
results on the remaining 78 streams (2,461 questions). On these
held-out streams, event lines improve accuracy by 2.0 points
$[0.5,3.6]$ on individual recordings ($p=0.023$) and 2.6 points
$[0.9,4.4]$ on stitched streams ($p=0.004$).

We also tested longer, 60-word lines and nine retrieval or prompt
variants. Only one variant improved development accuracy
($+4.2$ points), and that gain did not carry over to the held-out
streams. All ablations below use one \gpt{} run per condition,
evaluated on the questions answered by every condition in the
table. We use exact McNemar tests; overall accuracy varies by
about $\pm1$ point across repeated runs.

The same event lines were built on the \hdepic frames (36,633 lines, one per
sampled frame) and shown to the answerer as the 16 lines that best match the
question, together with the names of the retrieved candidates, since \hdepic
questions mark objects by box. Table~\ref{tab:hdepic-events} compares them with
the same memory re-answered in the same session. The mean moves by $+0.7$
$[-1.3,+2.5]$, no prototype changes significantly, and the gap to AMEGO on
fixture-interaction counting and stationary localization remains.

\begin{table}[htbp]
\centering
\small
\setlength{\tabcolsep}{4pt}
\caption{Event lines on \hdepic (2,400 questions, \gpt): accuracy (\%) per
prototype for the memory re-answered in the same session (control) and with event
lines; $\Delta$ is paired, with 95\% bootstrap intervals. AMEGO (adapted) is shown
for reference.}
\label{tab:hdepic-events}
\begin{tabular}{lrrrr}
\toprule
Prototype & Control & Event lines & $\Delta$ [95\% CI] & AMEGO \\
\midrule
Object location                & 48.8 & 50.8 & $+2.0$ $[-1.6,+5.6]$ & 39.8 \\
Object contents retrieval      & 30.5 & 35.0 & $+4.5$ $[-1.5,+10.5]$ & 22.0 \\
Fixture location               & 56.2 & 58.0 & $+1.8$ $[-1.0,+4.6]$ & 12.8 \\
Fixture interaction counting   & 41.3 & 40.7 & $-0.7$ $[-4.3,+3.0]$ & 47.0 \\
Movement itinerary             & 40.4 & 42.2 & $+1.8$ $[-2.4,+5.8]$ & 26.8 \\
Movement counting              & 42.5 & 43.0 & $+0.5$ $[-4.5,+5.5]$ & 34.5 \\
Stationary object localization & 36.0 & 32.0 & $-4.0$ $[-11.5,+3.5]$ & 55.5 \\
\midrule
Mean (prototype-averaged)      & 41.9 & 42.6 & $+0.7$ $[-1.3,+2.5]$ & 34.7 \\
\bottomrule
\end{tabular}
\end{table}

\subsection{Leave-one-out ablation}
\label{app:ucs-loo}

\begin{figure}[t]
\centering
\includegraphics[width=\textwidth]{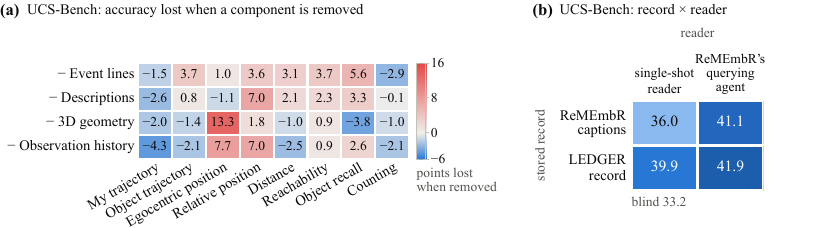}
\caption{\ucs stitched streams, \gpt. (a)~Accuracy lost on each question category
when one component is removed from the memory with event lines (single runs, 2,246
questions answered by every arm; Table~\ref{tab:ucs-loo}). (b)~The same two records
read by our single-shot reader and by ReMEmbR's querying agent (2,777 paired
questions; Table~\ref{tab:ucs-reader}).}
\label{fig:components-ucs}
\end{figure}

Table~\ref{tab:ucs-loo} removes one part of the memory with event lines at a
time (Figure~\ref{fig:components-ucs}a). No removal moves the overall score by more than 1.3 points on the questions
every arm answered, whereas the changes by category reach 13 points (egocentric
position without 3D geometry) and differ in sign across categories. On the larger
set of questions answered by both the reference and a given arm (3,103 to 3,171),
removing the event lines costs 2.2 points ($p=0.006$) and the descriptions 1.8
($p=0.02$); the other removals stay within noise.

\begin{table}[htbp]
\centering
\scriptsize
\setlength{\tabcolsep}{2.5pt}
\caption{Leave-one-out on the stitched streams (2,246 questions answered by every
arm). Category columns: my trajectory, object trajectory, egocentric
position, relative position, distance comparison, reachability, object recall,
counting.}
\label{tab:ucs-loo}
\begin{tabular}{lr rrrrrrrr}
\toprule
Memory & All & MyTraj & ObjTraj & EgoPos & RelPos & Dist & Reach & Recall & Count \\
$n$ & 2,246 & 348 & 138 & 195 & 329 & 198 & 218 & 268 & 552 \\
\midrule
Full (objects + event lines)     & 39.0 & 35.6 & 39.9 & 37.4 & 40.7 & 47.0 & 57.8 & 42.5 & 28.3 \\
$-$ event lines                  & 37.8 & 37.1 & 36.2 & 36.4 & 37.1 & 43.9 & 54.1 & 36.9 & 31.2 \\
$-$ object descriptions          & 37.6 & 38.2 & 39.1 & 38.5 & 33.7 & 44.9 & 55.5 & 39.2 & 28.4 \\
$-$ 3D geometry                  & 38.6 & 37.6 & 41.3 & 24.1 & 38.9 & 48.0 & 56.9 & 46.3 & 29.3 \\
$-$ observation history          & 38.4 & 39.9 & 42.0 & 29.7 & 33.7 & 49.5 & 56.9 & 39.9 & 30.4 \\
$-$ relations                    & 39.9 & 37.9 & 46.4 & 37.4 & 38.9 & 49.5 & 58.7 & 41.8 & 29.2 \\
$-$ places                       & 39.2 & 37.4 & 42.0 & 37.9 & 38.3 & 47.0 & 52.8 & 41.8 & 31.2 \\
$-$ objects (event lines only)   & 38.1 & 40.2 & 42.8 & 24.6 & 35.3 & 47.5 & 57.8 & 40.3 & 29.7 \\
\bottomrule
\end{tabular}
\end{table}

\subsection{Record against reader}
\label{app:ucs-reader}
ReMEmbR \citep{anwar2024remembr} captions every sampled frame with VILA, stores
the camera position and time of each caption, and answers with an agent that
issues embedding searches over the captions in several rounds, about four calls
and 23k tokens per question; our reader is one keyword lookup and one call, about
3 to 5k tokens. Table~\ref{tab:ucs-reader} crosses the two records with the two
readers (Figure~\ref{fig:components-ucs}b). To be indexed by ReMEmbR's embedder, our record is re-expressed as one
text item per sampled moment (event line, place, and the objects in view with
their descriptions and positions); adding each object's direction and distance
from the wearer to the items did not help. We use ReMEmbR's captions read by its own agent as the reference.
On paired questions, switching those captions to our reader lowers
accuracy by 5.0 points ($p<0.001$, exact McNemar test). Our records
score 1.2 points below the reference with our reader ($p=0.35$)
and 0.8 points above it with ReMEmbR's agent ($p=0.42$).

ReMEmbR's captions still outperform the blind control under our
reader by 2.6 points ($p=0.01$), showing that they contain useful
evidence. However, their advantage under the native-system comparison
does not persist when both memories use the same reader. Routing
spatial questions to our reader and all remaining questions to
ReMEmbR's agent yields 42.0\% accuracy on 3,357 questions.

\begin{table}[htbp]
\centering
\scriptsize
\setlength{\tabcolsep}{2.5pt}
\caption{Record against reader on the stitched streams (2,777 questions answered by
every arm). R1 is our single-shot reader; R2 is ReMEmbR's retrieval agent.
Blind on these streams is 33.2.}
\label{tab:ucs-reader}
\begin{tabular}{lr rrrrrrrr}
\toprule
Record $\rightarrow$ reader & All & MyTraj & ObjTraj & EgoPos & RelPos & Dist & Reach & Recall & Count \\
$n$ & 2,777 & 423 & 156 & 270 & 409 & 288 & 257 & 307 & 667 \\
\midrule
ReMEmbR captions $\rightarrow$ R1              & 36.0 & 38.5 & 37.8 & 29.6 & 32.3 & 41.7 & 55.3 & 32.9 & 30.6 \\
ReMEmbR captions $\rightarrow$ R2 (= ReMEmbR)  & 41.1 & 45.9 & 50.0 & 28.1 & 37.9 & 42.7 & 48.6 & 44.3 & 37.9 \\
Our event lines only $\rightarrow$ R1          & 38.2 & 40.2 & 39.7 & 26.7 & 33.5 & 46.2 & 58.0 & 41.4 & 31.8 \\
Our objects + ReMEmbR captions $\rightarrow$ R1 & 38.5 & 39.5 & 36.5 & 35.6 & 38.9 & 45.8 & 55.6 & 39.7 & 29.1 \\
\sysname (objects + event lines) $\rightarrow$ R1 & 39.9 & 37.1 & 40.4 & 36.7 & 40.6 & 45.8 & 58.8 & 45.6 & 30.0 \\
\sysname $\rightarrow$ R2                      & 41.9 & 49.2 & 47.4 & 30.4 & 35.9 & 45.1 & 54.5 & 44.6 & 36.7 \\
\sysname + egocentric geometry $\rightarrow$ R2 & 41.3 & 47.0 & 48.1 & 25.6 & 34.2 & 44.1 & 55.6 & 45.6 & 38.2 \\
\bottomrule
\end{tabular}
\end{table}

\subsection{Open answering models}
\label{app:ucs-answerers}
Table~\ref{tab:ucs-answerers} replaces \gpt by \qwen for our memory and its blind
control, and runs DirectMe in its published configuration, Qwen3-VL-8B reading the
graph with keyframes. On all 3,367 questions, with 58 unparsed \qwen answers
scored as wrong, the memory's gain over the \qwen blind control is 3.4 points
($p<0.001$).

\begin{table}[htbp]
\centering
\scriptsize
\setlength{\tabcolsep}{2.5pt}
\caption{Answering models on the stitched streams (3,300 questions answered by
every arm).}
\label{tab:ucs-answerers}
\begin{tabular}{lr rrrrrrrr}
\toprule
Arm & All & MyTraj & ObjTraj & EgoPos & RelPos & Dist & Reach & Recall & Count \\
$n$ & 3,300 & 511 & 191 & 305 & 480 & 318 & 308 & 374 & 813 \\
\midrule
\qwen, blind                              & 32.3 & 39.5 & 46.6 & 22.3 & 28.5 & 38.1 & 49.0 & 34.2 & 20.9 \\
\sysname + event lines, \qwen             & 36.2 & 38.0 & 35.6 & 36.1 & 34.8 & 42.8 & 53.6 & 38.0 & 26.3 \\
DirectMe graph + keyframes, Qwen3-VL-8B   & 28.0 & 33.3 & 33.5 & 24.9 & 27.1 & 36.5 & 46.8 & 26.2 & 15.5 \\
\sysname + event lines, \gpt              & 39.5 & 36.2 & 38.7 & 37.0 & 39.8 & 45.9 & 58.1 & 44.7 & 30.6 \\
ReMEmbR, \gpt                             & 40.8 & 45.0 & 50.3 & 27.5 & 38.5 & 42.5 & 49.4 & 43.6 & 37.0 \\
\bottomrule
\end{tabular}
\end{table}

\subsection{Delayed questions}
\label{app:delayed}
The 763 questions about an earlier scene that do not depend on the present
moment (object recall, counting, and object and wearer trajectories) are re-asked
at the end of the stream, about their original moment, from the memory held at
the end. Rewording a question as delayed moves the blind prior by $+5.5$ on the
same questions, so Table~\ref{tab:delayed} reports each arm's change net of
blind: the frame reader loses most, since its 48 frames must cover the whole
stream, and the memories lose 3 to 4 points. When the original moment is given
exactly, rounded to 1, 5, or 10 minutes, or not at all, ReMEmbR is insensitive to
the hint (40.9 to 42.7), \sysnameagent is within 1 to 2
points of it at every level (n.s.), and our single-shot reader is weaker (30.7 to
36.7) because its time bonus favours what was seen near the question, which for a
delayed question is the end of the stream.

\begin{table}[htbp]
\centering
\small
\caption{Delayed questions (763): accuracy at the question's own moment and
when re-asked at the end of the stream, with the change net of the blind control.}
\label{tab:delayed}
\begin{tabular}{lrrrr}
\toprule
Evidence & At its moment & Delayed & $\Delta$ & $\Delta$ net of blind \\
\midrule
Blind                           & 32.4 & 37.9 & $+5.5$ & -- \\
48 frames                       & 44.0 & 39.6 & $-4.4$ & $-9.9$ \\
DirectMe graph                  & 25.9 & 28.3 & $+2.5$ & $-3.0$ \\
ReMEmbR                         & 40.6 & 42.7 & $+2.1$ & $-3.4$ \\
\sysname, objects only          & 34.1 & 32.8 & $-1.4$ & $-6.9$ \\
\sysname + event lines          & 35.1 & 36.6 & $+1.4$ & $-4.1$ \\
\bottomrule
\end{tabular}
\end{table}

\FloatBarrier

%% file: sections/appendix_length.tex
\section{Accuracy by Recording Length}
\label{app:length}

Figure~\ref{fig:length} summarizes this appendix. Every arm is answered by \gpt
and paired on the same questions, and a question is kept only if every arm in
the table answered it; an invalid or missing answer counts as wrong. Accuracies
here are over questions, so the \hdepic ``All'' row (44.6) differs from the
prototype-averaged score of Table~\ref{tab:hdepic-main} (42.6). Lengths are the
durations of the source videos. Intervals are Wilson intervals for accuracies and
paired bootstrap intervals over videos (2,000 draws) for differences; the trend
is a logistic regression of correctness on $\log_2$ of the recording length, so
its slope is the change in log-odds per doubling. \sysnameagent is our memory
read by ReMEmbR's querying agent (Appendix~\ref{app:ucs-reader}); the ``agent''
columns below are this arm.

\paragraph{\hdepic.} All 2,400 questions on 150 videos
(Table~\ref{tab:length-hdepic}). Every arm loses accuracy on longer recordings,
and the blind control loses it fastest (Table~\ref{tab:length-trend}), so the
questions on long recordings are harder to guess. Our memory's gain over the blind
control is $+9.9$ $[+2.5,+17.7]$, $+18.0$, $+16.2$, $+14.4$, and $+14.8$
$[+5.9,+23.8]$ points across the five bins, and its lead over ReMEmbR is 12 to 19
points in every bin. Length is partly a proxy for participant, since each
participant records one kitchen, but every bin holds six to nine of the nine
participants.

\begin{table}[htbp]
\centering
\small
\setlength{\tabcolsep}{3pt}
\caption{\hdepic accuracy (\%, over questions) by recording length (minutes).}
\label{tab:length-hdepic}
\begin{tabular}{lrr ccccccc}
\toprule
 & & & \multicolumn{2}{c}{\sysname} & & & & & \\
\cmidrule(lr){4-5}
Length & Videos & Questions & single-shot & agent & ReMEmbR & AMEGO & DirectMe & OSNOM & Blind \\
\midrule
$<$5 & 45 & 284 & \textbf{48.6} & 35.6 & 29.6 & 35.2 & 33.8 & 38.4 & 38.7 \\
5--10 & 22 & 266 & \textbf{48.5} & 39.1 & 32.7 & 31.6 & 27.1 & 32.7 & 30.5 \\
10--20 & 37 & 769 & \textbf{45.1} & 33.2 & 27.8 & 31.3 & 29.6 & 31.1 & 28.9 \\
20--40 & 34 & 710 & \textbf{42.4} & 30.1 & 23.4 & 31.1 & 30.3 & 32.7 & 28.0 \\
40--75 & 12 & 371 & \textbf{41.8} & 37.2 & 29.9 & 31.3 & 26.4 & 30.5 & 27.0 \\
\midrule
All & 150 & 2,400 & \textbf{44.6} & 33.8 & 27.6 & 31.8 & 29.5 & 32.5 & 29.7 \\
\bottomrule
\end{tabular}
\end{table}

\paragraph{\ucs.} The 3,264 questions of the single-recording arm of the stitched
experiment, on 206 recordings of up to 20 minutes (Table~\ref{tab:length-ucs}).
No memory changes with length, and neither does the blind control; the 48-frame
reader, the only arm that sees video, loses accuracy with every doubling of
length. The bins are not balanced across sources (the 11--13-minute bin is mostly
EgoLife and holds 63\% of the questions), so we repeat the fit within
EPIC-Kitchens, the one source present in every bin (45 recordings, 801
questions): the frame reader still declines ($p=0.05$) and no memory does
(Table~\ref{tab:length-trend}).

\begin{table}[htbp]
\centering
\small
\setlength{\tabcolsep}{3pt}
\caption{\ucs accuracy (\%) by recording length (minutes), single recordings.}
\label{tab:length-ucs}
\begin{tabular}{lrr cccccc}
\toprule
 & & & \multicolumn{2}{c}{\sysname} & & & & \\
\cmidrule(lr){4-5}
Length & Videos & Questions & single-shot & agent & ReMEmbR & DirectMe & Blind & \vidsq 48 frames \\
\midrule
$<$2 & 16 & 236 & 36.0 & 41.5 & 37.7 & 31.4 & 33.1 & 61.4 \\
2--6 & 23 & 370 & 38.1 & 43.5 & 42.4 & 28.6 & 32.4 & 50.0 \\
6--11 & 18 & 321 & 45.8 & 49.5 & 46.1 & 29.9 & 34.9 & 55.1 \\
11--13 & 123 & 2,050 & 38.6 & 40.4 & 40.8 & 29.3 & 32.6 & 44.5 \\
13--20 & 26 & 287 & 37.3 & 44.3 & 41.5 & 36.2 & 35.2 & 48.1 \\
\midrule
All & 206 & 3,264 & 39.0 & 42.1 & 41.4 & 30.1 & 33.1 & 47.7 \\
\bottomrule
\end{tabular}
\end{table}

\paragraph{\vqthreed.} The validation clips come in two lengths, about 5 and
about 8 minutes, so this is not a length sweep (Table~\ref{tab:length-vq3d}). The
measure is the share of all 164 queries localized within 1\,m, with an
unanswered query counted as a miss.

\begin{table}[htbp]
\centering
\small
\setlength{\tabcolsep}{3pt}
\caption{\vqthreed: queries localized within 1\,m (\%) by clip length (minutes).}
\label{tab:length-vq3d}
\begin{tabular}{lrr cccccc}
\toprule
 & & & \multicolumn{2}{c}{\sysname} & & & & \\
\cmidrule(lr){4-5}
Length & Clips & Queries & single-shot & agent & ReMEmbR & DirectMe & OSNOM & Clip centroid \\
\midrule
$\approx$5 & 16 & 66 & 56.1 & 59.1 & 22.7 & 10.6 & 34.8 & 7.6 \\
$\approx$8 & 26 & 93 & 43.0 & 45.2 & 23.7 & 21.5 & 30.1 & 11.8 \\
16 & 2 & 5 & 20.0 & 0.0 & 20.0 & 40.0 & 0.0 & 20.0 \\
\midrule
All & 44 & 164 & 47.6 & 49.4 & 23.2 & 17.7 & 31.1 & 10.4 \\
\bottomrule
\end{tabular}
\end{table}

\begin{table}[htbp]
\centering
\small
\setlength{\tabcolsep}{4pt}
\caption{Change in the log-odds of a correct answer per doubling of recording
length, with its $p$-value in parentheses.}
\label{tab:length-trend}
\begin{tabular}{lccc}
\toprule
Arm & \hdepic & \ucs & \ucs, EPIC-Kitchens only \\
\midrule
\sysname, single-shot reader & $-0.072$ (0.028) & $+0.013$ (0.69) & $+0.021$ (0.74) \\
\sysnameagent & $-0.054$ (0.12) & $-0.007$ (0.82) & $+0.022$ (0.72) \\
ReMEmbR & $-0.060$ (0.099) & $+0.016$ (0.63) & $+0.086$ (0.17) \\
DirectMe & $-0.072$ (0.044) & $+0.013$ (0.71) & $+0.010$ (0.89) \\
AMEGO (adapted) & $-0.076$ (0.029) & -- & -- \\
OSNOM (adapted) & $-0.084$ (0.016) & -- & -- \\
Blind & $-0.155$ ($<$10$^{-4}$) & $+0.008$ (0.82) & $-0.066$ (0.3) \\
\vidsq 48 frames & -- & $-0.157$ ($<$10$^{-4}$) & $-0.122$ (0.05) \\
\bottomrule
\end{tabular}
\end{table}
\FloatBarrier

%% file: sections/appendix_analysis.tex
\section{Failure Analysis and Qualitative Example}
\label{app:analysis}

\subsection{When an answer is wrong, where was it lost?}
\label{app:loss}

Table~\ref{tab:loss} follows the evidence through the pipeline. The largest
measured losses happen before the answerer reads the record: on development
videos, giving \gpt the ground-truth trajectory of the queried object in place of
the stored one adds 12 points, and naming the fixtures it rested on adds another
25, whereas a fixed rule improves on the answerer's reasoning by only 3 points on
fixture location (Appendices~\ref{app:hdepic-geometry} and~\ref{app:implementation}). Each weak category of
RQ1 (Section~\ref{sec:evaluation}) maps to one row: stationary localization and
fixture-interaction counting to what the record keeps about rest and hands,
counting to association, and long recordings to construction.

\begin{table}[htbp]
\centering
\footnotesize
\setlength{\tabcolsep}{3pt}
\renewcommand{\arraystretch}{1.1}
\caption{Where evidence is lost, by pipeline stage: what each stage loses, the
questions it affects, and the measurement behind it.}
\label{tab:loss}
\begin{tabular}{@{}>{\raggedright\arraybackslash}p{0.13\textwidth}>{\raggedright\arraybackslash}p{0.27\textwidth}>{\raggedright\arraybackslash}p{0.22\textwidth}>{\raggedright\arraybackslash}p{0.32\textwidth}@{}}
\toprule
Stage & What is lost & Questions affected & Evidence \\
\midrule
Sampling (one frame per 4\,s) & objects seen only briefly; views for triangulation & all & 2\,s sampling did not help a ten-video pilot (App.~\ref{app:hdepic-design}) \\
Tagging & objects outside the per-video vocabulary never enter memory & any about those objects & the best tag set misses about one \vqthreed query object in ten (Table~\ref{tab:vq3d-tags}) \\
3D lifting & depth along the viewing ray & metric localization & the error lies almost entirely along the ray (App.~\ref{app:hdepic-geometry}) \\
Association & one object split into several tracks & counting & the track count is exact for 11\% of \ucs count questions (App.~\ref{app:ucs-counting}) \\
Record content & continuous rest intervals; hand events & stationary localization, fixture-interaction counting & rest-segment starts are near chance; hand tracklets score higher (App.~\ref{app:hdepic-motion}, Table~\ref{tab:hdepic-membase-proto}) \\
Construction & labels and metric scale at a scene cut; tractability with length & after a scene change; hour-long recordings & RQ5; 21 of 167 HourVideo memories built (App.~\ref{app:ucs}) \\
Retrieval & a wrong instance of the right category & the \vqthreed error tail & 10\% of queries are off by more than 3\,m (App.~\ref{app:vq3d-published}) \\
Reasoning & retrieved evidence left unused & fixture location & a fixed rule beats \gpt on the same evidence (App.~\ref{app:implementation}) \\
\bottomrule
\end{tabular}
\end{table}

\subsection{Three memories on one stream}
\label{app:qualitative}

\input{sections/qualitative_figure.tex}

Figure~\ref{fig:qualitative} shows DirectMe, ReMEmbR, and \sysname answering from
their own memories of one stitched stream of three EPIC-Kitchens recordings, and
localizing a chair in a separate \vqthreed clip. The questions were chosen to show
how the memories differ, not how often; rates are in Tables~\ref{tab:ucs-main}
and~\ref{tab:stitch}. Our memory answers all four stream questions, DirectMe's
graph answers one, and ReMEmbR none.

The differences follow from what each memory keeps. For Q2, ``What did I take out
of the cabinet?'', DirectMe's graph has no node for the object, and ReMEmbR's
captions near that time do not name it. Our detector labelled the object a coffee
maker, but its segment description reads ``a digital immersion circulator, not a
coffee maker \ldots\ the person's hands are holding the device'', and the answerer
matches it to the heating rod. For the chair in Q5, our stored position is
0.73\,m from the annotated one, against 1.81\,m for DirectMe and 2.37\,m for
ReMEmbR, which stores where the wearer was rather than where the object is.
\FloatBarrier

%% file: sections/qualitative_figure.tex
\begin{figure}[t]
\centering
\includegraphics[width=\textwidth]{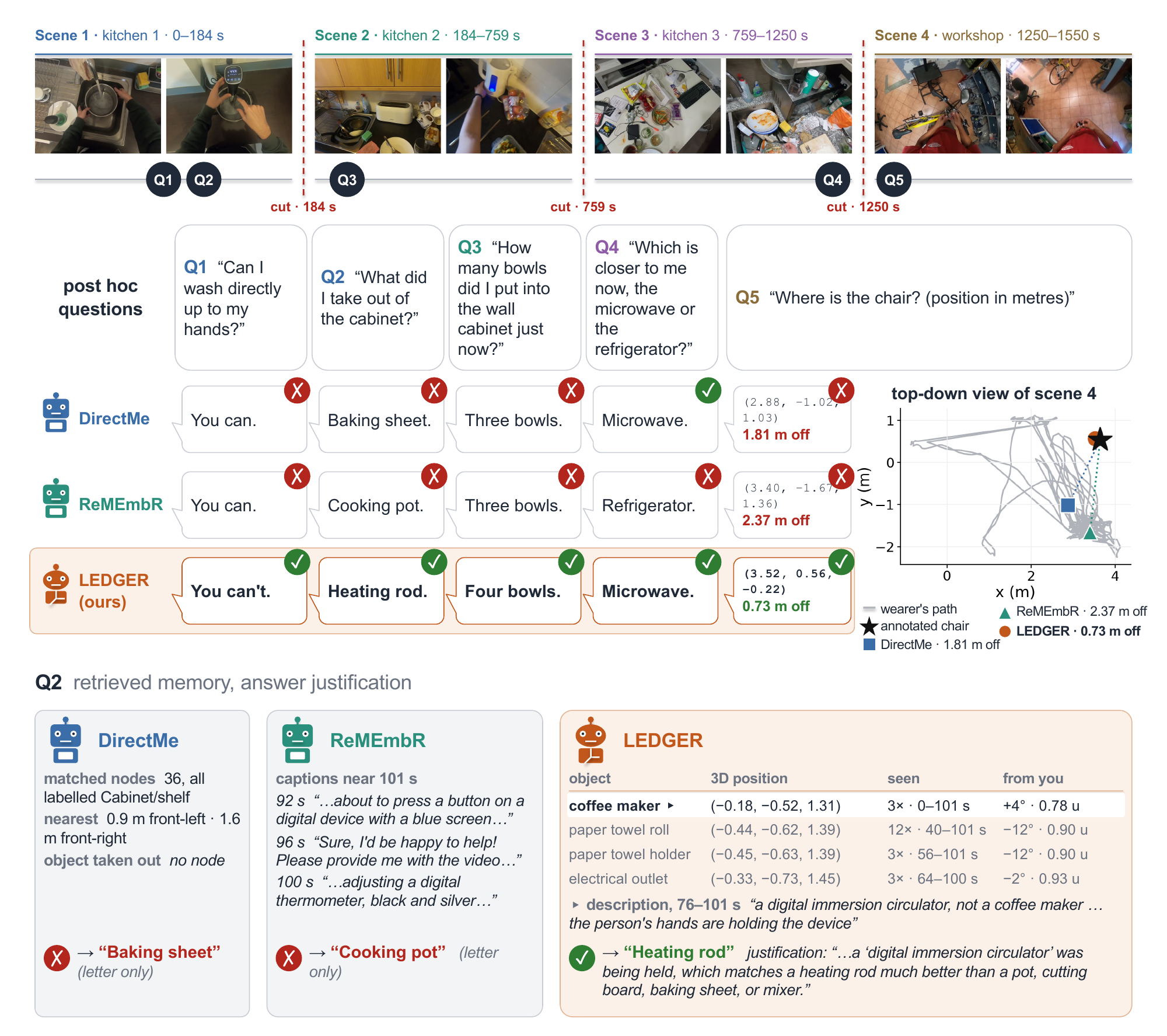}
\caption{Three memories answer questions about a multi-scene egocentric stream
from their stored records alone. Scenes 1--3 form one \ucs stream that joins
three EPIC-Kitchens recordings; DirectMe, ReMEmbR, and \sysname each build one
memory over the stream and answer Q1--Q4 from it as text, without frames. Scene 4
is a separate \egofour \vqthreed clip placed on the same timeline for
illustration, with memories built from that clip alone; Q5 asks for the chair's
position in metres, and the map (right) shows each memory's answer.
\textbf{Bottom:} the evidence each memory retrieves for Q2. \sysname's detector
labelled the object a coffee maker, but the segment description records what the
wearer's hands were holding; DirectMe's graph has no node for the object, and
ReMEmbR's captions near that time do not name it. Questions were selected for
illustration; aggregate results are in Tables~\ref{tab:ucs-main}
and~\ref{tab:stitch}.}
\label{fig:qualitative}
\end{figure}

%% file: sections/appendix_cost.tex
\section{Answer-Time Token Cost}
\label{app:cost}

A memory is built once per recording and read many times, whereas frames are
re-read at every question. This appendix measures what each memory costs the
answering model at answer time, on the same questions for every arm. Tokens are
the answering model's own usage, averaged per question, and calls are
language-model calls per question. Memory construction is excluded: every method
builds its memory locally, with no API calls, so its cost is compute time rather
than tokens. Published video models and video-retrieval methods that we did not run
on these questions are absent.

\begin{table}[htbp]
\centering
\small
\setlength{\tabcolsep}{5pt}
\caption{\novidsq Answer-time cost on \hdepic (2,400 questions, \gpt unless
stated). Accuracy is prototype-averaged, as in Table~\ref{tab:hdepic-main}. No
arm here reads frames.}
\label{tab:cost-hdepic}
\begin{tabular}{llrrrr}
\toprule
Memory & Read at answer time & Acc.\ (\%) & Input tok & Output tok & Calls \\
\midrule
Blind                    & question and options only      & 29.7 & 412    & 184 & 1.0 \\
ReMEmbR                  & captions, retrieval agent       & 29.8 & 25,201 & 207 & 4.1 \\
OSNOM (adapted)          & tracked object map              & 30.2 & 8,762  & 111 & 1.0 \\
DirectMe                 & scene graph, its own prompt     & 30.7 & 2,362  & 5   & 1.0 \\
\sysname, \qwen          & object memory                   & 33.6 & 1,938  & 381 & 1.0 \\
AMEGO (adapted)          & hand--object interaction log    & 34.7 & 792    & 124 & 1.0 \\
\sysname                 & object memory                   & \textbf{42.6} & 1,750 & 231 & 1.0 \\
\sysnameagent           & object memory, querying agent   & 34.8 & 22,850 & 199 & 3.6 \\
\bottomrule
\end{tabular}
\end{table}

On \hdepic (Table~\ref{tab:cost-hdepic}), the stored record is read in one call of about 1.8k input tokens, and
the two memories that cost less are the two that keep less: AMEGO's interaction
log at 0.8k admits only handled objects, and the blind control at 0.4k stores
nothing. The memories that cost more do not answer better here. ReMEmbR reads
about 14 times as many input tokens over roughly four calls and scores at the
blind level, because its captions carry no object positions.

\begin{table}[htbp]
\centering
\small
\setlength{\tabcolsep}{4pt}
\caption{Answer-time cost on \ucs (2,772 EgoLife and TeleEgo questions, \gpt).
Green marks arms whose answerer sees no frames, red those that do. Frame arms
read 48 uniformly sampled frames up to the question time. This set has one
question more than Table~\ref{tab:ucs-main}, so DirectMe's memory-only accuracy
reads 29.9 here against 30.0 there.}
\label{tab:cost-ucs}
\begin{tabular}{llrrrr}
\toprule
Evidence & Read at answer time & Acc.\ (\%) & Input tok & Output tok & Calls \\
\midrule
\novidsq Blind             & question and options only     & 33.8 & 452    & 149 & 1.0 \\
\novidsq DirectMe          & scene graph                   & 29.9 & 1,937  & 2   & 1.0 \\
\novidsq \sysname          & object memory                 & 38.5 & 2,044  & 211 & 1.0 \\
\midrule
\vidsq Frames              & 48 frames                     & 44.4 & 9,615  & 50  & 1.0 \\
\vidsq Frames, DM routed   & frames; graph on spatial q.   & 42.9 & 10,406 & 49  & 1.0 \\
\vidsq Frames, DM own      & frames and graph always       & 39.8 & 11,377 & 48  & 1.0 \\
\vidsq Frames, \sysname    & frames; memory on spatial q.  & \textbf{46.3} & 10,529 & 137 & 1.0 \\
\bottomrule
\end{tabular}
\end{table}

The \ucs arms (Table~\ref{tab:cost-ucs}) separate the two costs. Reading 48 frames costs about 9.6k input
tokens per question, nearly five times the 2.0k that reading our record costs, and it
buys 5.9 points over the record alone. Adding the record to the frames costs a
further 0.9k, about a tenth of the frame budget, and recovers 1.9 points on top
of frames. A memory therefore substitutes for frames at roughly a fifth of the
answer-time cost, and complements them at a small fraction of it.

\paragraph{Against ReMEmbR on the questions it answered.}
ReMEmbR returns an answer for 2,183 of the 2,772 questions. Restricted to those,
it scores 41.3\% against 38.5\% for our memory, so it is ahead on these single
recordings, and it reads 18,917 input tokens over 3.7 calls per question against
2,041 in one call, about nine times the input for one call in place of four. The
blind control scores 33.3\% on the same subset. Whether that accuracy is worth
the cost depends on the budget; the comparison of what each record preserves,
independent of how it is read, is in Appendix~\ref{app:ucs-reader}.